\documentclass[lettersize,journal]{IEEEtran}
\usepackage{amsmath,amsfonts}
\usepackage{algorithmic}
\usepackage{algorithm}
\usepackage{array}
\usepackage[caption=false,font=normalsize,labelfont=sf,textfont=sf]{subfig}
\usepackage{textcomp}
\usepackage{stfloats}
\usepackage{url}
\usepackage{verbatim}
\usepackage{graphicx}
\usepackage{cite}
\usepackage[utf8]{inputenc} % allow utf-8 input
\usepackage[T1]{fontenc}    % use 8-bit T1 fonts
\usepackage{hyperref}       % hyperlinks
\usepackage{url}            % simple URL typesetting
\usepackage{booktabs}       % professional-quality tables
\usepackage{amsfonts}       % blackboard math symbols
\usepackage{nicefrac}       % compact symbols for 1/2, etc.
\usepackage{microtype}      % microtypography
\usepackage{xcolor}         % colors
\usepackage{graphicx} 
\usepackage{multirow}
\usepackage{amsmath}
\usepackage{colortbl}
\usepackage{caption}
\begin{document}

\title{MZ-Rain: Moisture-Budget-Guided Zero-Inflated Model for Station-Level Precipitation Nowcasting}

\author{
Yifang~Zhang,
Shengwu~Xiong,
Henan~Wang,
Wenjie~Yin,
Yuqiang~Zhang,
Chen~Zhou,
Hua~Chen,
Qile~Zhao,
and~Pengfei~Duan% <-this % stops a space
\thanks{Yifang~Zhang, Shengwu~Xiong, Henan~Wang and Pengfei~Duan are with the Sanya Science and Education Innovation Park, Wuhan University of Technology, Sanya, 572000, China and also with the School of Computer Science and Artificial Intelligence, Wuhan University of Technology, Wuhan 430070, China (e-mail: yifangzhang@whut.edu.cn; xiongsw@whut.edu.cn; 361332@whut.edu.cn; duanpf@whut.edu.cn).}%
\thanks{Wenjie~Yin, Yuqiang~Zhang, and Chen~Zhou are with the School of Earth and Space Science and Technology, Wuhan University, Wuhan 430072, China (e-mail: windsoryin@whu.edu.cn; 
yqzhang\_3@whu.edu.cn;
chenzhou@whu.edu.cn).}%
\thanks{Hua~Chen is with the School of Water Resources and Hydropower Engineering, Wuhan University, Wuhan 430062, China (e-mail: chua@whu.edu.cn).}%
\thanks{Qile~Zhao is with the GNSS Research Center, Wuhan University, Wuhan 430062, China (e-mail: zhaoql@whu.edu.cn).}%
\thanks{Corresponding author: Pengfei Duan (duanpf@whut.edu.cn).}

}

\markboth{Journal of \LaTeX\ Class Files,~Vol.~14, No.~8, August~2021}%
{Shell \MakeLowercase{\textit{et al.}}: A Sample Article Using IEEEtran.cls for IEEE Journals}


\maketitle

\begin{abstract}
Accurate station-level precipitation nowcasting is critical for agriculture, water resource management, and disaster prevention, which typically is formulated
as a time series forecasting problem. However, conventional time-series modeling techniques face two major challenges in addressing station-level precipitation nowcasting: (1) \textbf{Lack of Physics-Guided Modeling}, where meteorological variables are treated as a homogeneous set without accounting for their distinct roles in precipitation formation, leads to predictions that deviate from the physical processes governing precipitation. (2) \textbf{Severe zero inflation in precipitation}, where dry intervals dominate the dataset, obscuring meaningful precipitation patterns and complicating the predictive modeling. To address these challenges, we propose \textbf{MZ-Rain}, a moisture-budget-guided zero-inflated sLSTM framework for station-level precipitation nowcasting. Guided by the moisture budget equation, MZ-Rain decomposes the precipitation formation process into process-specific pathways corresponding to moisture storage, moisture transport, surface evaporation, and precipitation persistence, and captures their temporal evolution through dedicated sLSTM branches. To account for the zero-inflated nature of precipitation, MZ-Rain introduces an adaptive Tweedie modeling strategy that adaptively modulates the rainfall mean while jointly learning precipitation occurrence as an auxiliary task, enabling the model to better balance dry-wet discrimination and quantitative precipitation estimation. Extensive experiments across diverse geographical and climatic regimes demonstrate that MZ-Rain consistently outperforms strong baselines on multiple evaluation metrics, including CSI, FAR, MSE, and MAE. In particular, the model exhibits superior skill in forecasting heavy precipitation events, while benefiting from physically grounded process modeling. Code and datasets are available at \url{https://anonymous.4open.science/r/MZ-Rain-7888}. 
\end{abstract}


\section{Introduction}
Accurately predicting precipitation within the next 0–6 hours, commonly referred to as precipitation nowcasting, is of great importance for disaster prevention, flood control, and real-time operational decision-making \cite{zhang2023skilful,nearing2024global}. Within this forecasting horizon, station-level precipitation nowcasting aims to produce site-specific precipitation forecasts from historical precipitation, precipitable water vapor (PWV), and auxiliary meteorological observations \cite{9351749, 11424616,yin2025accurate}.
Unlike spatial nowcasting methods that predict the evolution of precipitation fields, station-level precipitation nowcasting focuses on localized rainfall evolution at individual observation sites, making it particularly relevant to location-specific applications such as urban waterlogging hotspots and critical hydraulic facilities. Benefiting from their strong capacity for temporal dependency modeling, deep learning methods have recently shown promising performance in station-level precipitation forecasting\cite{11424616,11367701,9351749}. 

Early work employed supervised learning methods such as SVM to improve hourly rainfall forecasting by leveraging temporal autocorrelation \cite{9351749}. Hybrid frameworks further combined rainfall event prediction and amount estimation, integrating LSTM with regression-based models to better handle different rainfall regimes \cite{11367701}. Zhang et al. address the dominance of zero values in precipitation sequences by introducing BFPF modules that explicitly enhance the representation of sparse rainfall signals \cite{11424616}. DET enhances Transformers with distribution-aware output modeling and tailored mechanisms for rare and extreme rainfall events \cite{Gao_Chen_Du_Yu_Bernal_Xu_2026}. 

Despite these advances, most existing approaches remain largely data-driven and do not explicitly account for the physical and statistical characteristics of precipitation. On one hand, due to the insufficient physics-guided modeling, conventional time-series models primarily learn statistical dependencies among meteorological observations, while heterogeneous atmospheric processes associated with moisture storage, transport, and surface moisture supply are often entangled within a shared representation. This entanglement obscures the distinct roles of these processes and limits the model's ability to capture their heterogeneous contributions to precipitation evolution.  
On the other hand, precipitation additionally exhibits severe zero inflation. Frequent dry periods dominate precipitation records, whereas wet and heavy-rain events are sparse but critical for practical forecasting. This imbalance biases conventional regression toward conservative near-zero predictions and creates a tension between suppressing false rainfall during dry periods and preserving accurate precipitation amounts under wet conditions.
These challenges highlight the need for a forecasting framework that simultaneously incorporates physics-guided modeling of precipitation processes and explicitly addresses the learning imbalance induced by the zero-inflated nature of precipitation.

\begin{figure}[h!]
\centering
\includegraphics[width=0.93\columnwidth]{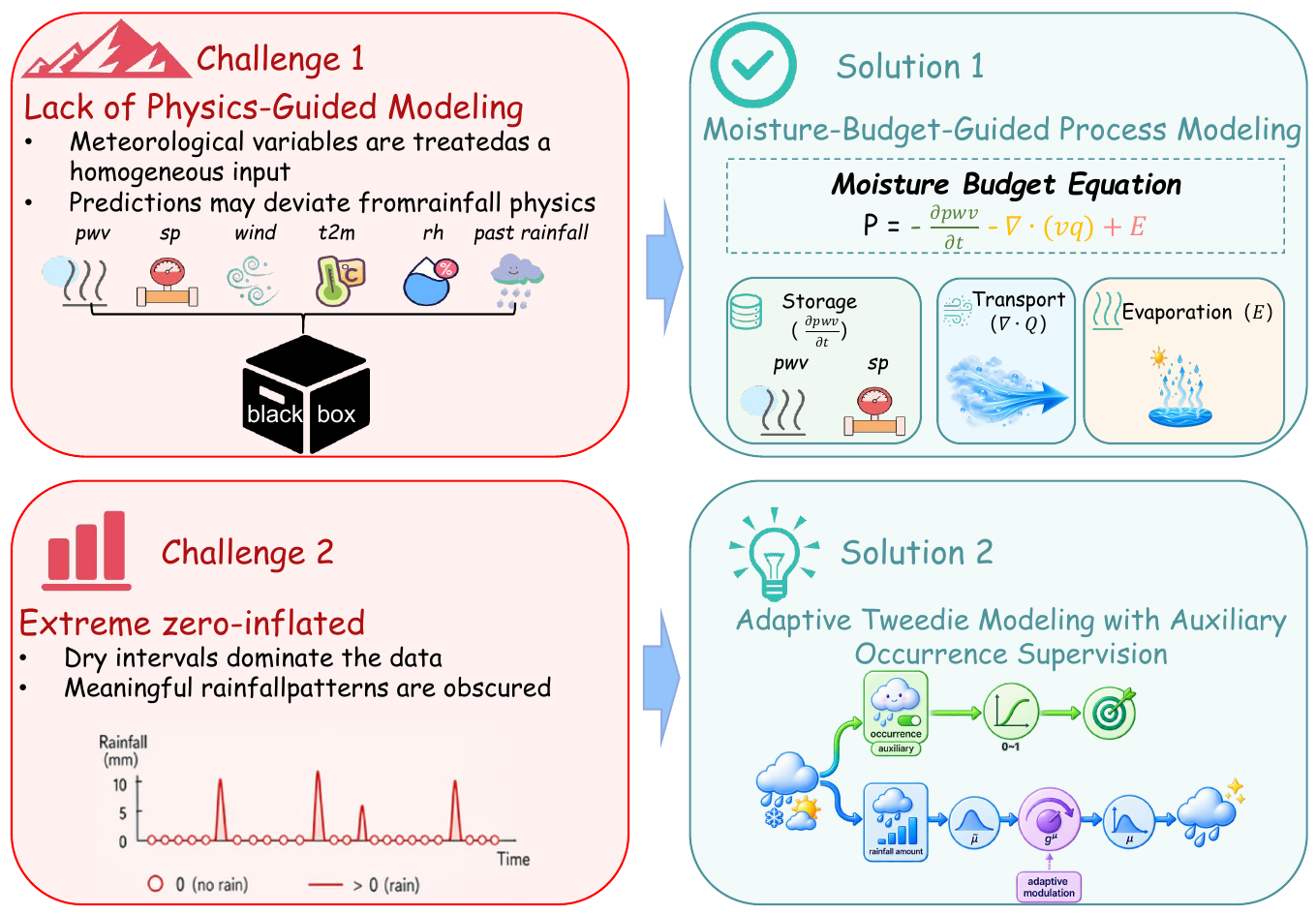}
\caption{Motivation of the proposed MZ-Rain, highlighting the two key challenges and their corresponding solutions.}
\label{fig:motivation}
\end{figure}

Motivated by these challenges, we propose \textbf{MZ-Rain}, a physics-guided framework for station-level precipitation nowcasting that integrates moisture-budget-guided process modeling with adaptive Tweedie modeling of zero-inflated precipitation, as illustrated in Fig.~\ref{fig:motivation}.

To address the lack of physics-guided modeling, MZ-Rain leverages the moisture budget equation as a structural prior \cite{cheng2022analysis} :
$
P_t = - \frac{\partial PWV_t}{\partial t} - \nabla \cdot\mathbf{Q}_t + E_t,
$
where $PWV_t$ denotes precipitable water vapor, $\mathbf{Q}_t$ is the vertically integrated horizontal moisture flux, and $E_t$ represents surface evaporation. These terms characterize three key moisture processes associated with precipitation evolution: atmospheric moisture storage, horizontal moisture transport and convergence, and surface moisture supply. Rather than imposing the moisture budget as a hard algebraic constraint, MZ-Rain uses its process structure to determine which atmospheric quantities are diagnosed and how they are organized for temporal representation learning.
Accordingly, MZ-Rain structures precipitation dynamics around three budget-related pathways: a \textbf{moisture storage pathway} characterized by PWV and its temporal evolution, a \textbf{moisture transport and convergence pathway} driven by diagnosed vertically integrated moisture-flux convergence, and a \textbf{surface evaporation pathway} driven by the surface moisture flux. Although the forecasting target is defined at an individual station, moisture-flux convergence is diagnosed from the spatially distributed moisture-flux field surrounding the station before being supplied to the temporal model. In addition, historical precipitation is modeled through a separate \textbf{precipitation persistence pathway} to capture short-term event continuity. Dedicated sLSTM branches \cite{beck2024xlstm} are employed to learn the heterogeneous temporal dynamics associated with these process-specific signals. In this way, the moisture budget serves as a physically motivated structural prior for representation learning rather than an exact closure condition on the neural prediction.

To address the extreme zero-inflation issue, MZ-Rain further introduces an adaptive Tweedie modeling strategy together with auxiliary occurrence supervision. 
The Tweedie-based formulation accommodates the coexistence of frequent zero precipitation and continuous, highly skewed positive amounts, while an adaptive calibration mechanism regulates quantitative precipitation estimation according to the learned forecasting state. A separate occurrence head provides explicit wet-dry supervision during training but does not directly enter the rainfall-amount prediction. This separation allows rainfall-event discrimination and quantitative precipitation estimation to provide complementary learning signals while preserving their distinct statistical roles. 

To evaluate the effectiveness of the proposed MZ-Rain model, we conducted experiments on six stations covering diverse geographical and climatic regions, using CSI, FAR, MSE, and MAE as evaluation metrics. Experiments across six geographically and climatically diverse stations show that MZ-Rain consistently improves both rainfall-event discrimination and quantitative precipitation estimation over strong baselines. Controlled analyses further demonstrate that moisture-budget-guided process modeling and adaptive Tweedie modeling provide complementary gains across forecast horizons and precipitation regimes. MZ-Rain also maintains robust performance under heavy-rain conditions.

Our main contributions are summarized as follows:
\begin{itemize}
\item
\textbf{Moisture-Budget-Guided Process Modeling for Precipitation Forecasting:} 
We introduce a physics-structured temporal framework that organizes precipitation-related atmospheric information into dedicated pathways for moisture storage, moisture transport and convergence, surface evaporation, and precipitation persistence under the guidance of the moisture budget equation.

\item
\textbf{Adaptive Tweedie Modeling with Auxiliary Occurrence Supervision:}
We introduce an adaptive Tweedie modeling strategy for quantitative precipitation estimation and retain precipitation occurrence as a separate auxiliary task. This formulation allows wet-dry discrimination and rainfall-amount estimation to provide complementary supervision without assigning the occurrence probability a direct multiplicative role in the final precipitation prediction.

\item
\textbf{Comprehensive Evaluation and Component Analysis:} We evaluate MZ-Rain across six geographically and climatically diverse stations and systematically analyze the contributions of diagnosed moisture-budget variables, process-specific pathway organization, adaptive Tweedie modeling, and auxiliary occurrence supervision across multiple forecast horizons and precipitation regimes.
\end{itemize}

\section{Related Work}
\subsection{Data-Driven Precipitation Forecasting}
Existing deep learning-based precipitation forecasting methods mainly improve prediction performance by modeling temporal dependencies, spatial structures, or sparse rainfall signals. For example,
Harilal et al.\cite{THOTTUNGALHARILAL2024108581}, using meteorological data from 1981 to 2023 across four regions in the UK, employed hybrid deep learning models (CNN-LSTM and RNN-LSTM) for daily precipitation prediction. The experimental results demonstrated that these hybrid models outperformed traditional LSTM and its variants. Yin et al.\cite{yin2025accurate}, based on the Informer\cite{zhou2021informer} model, integrated GNSS PWV data and ERA5 meteorological data, effectively capturing long-range dependencies in time series data. Zhang et al. \cite{11424616} address the dominance of zero values in precipitation sequences by introducing BFPF modules that explicitly enhance the representation of sparse rainfall signals. Geng et al.\cite{geng2023lstmatu} introduced the LSTMAtU-Net model, which combines LSTM units with a U-Net structure and incorporates Efficient Channel and Spatial Attention (ECSA) modules. This integration significantly enhanced the model's ability to capture long-term dependencies and spatial features, improving performance across various precipitation thresholds in precipitation nowcasting.

Although these studies have improved precipitation forecasting by enhancing temporal dependency modeling, spatial feature extraction, or sparse rainfall representation, most existing methods still treat meteorological variables as homogeneous predictors and rely primarily on data-driven feature learning. They rarely organize meteorological information according to the physical processes associated with precipitation formation. In contrast, our proposed MZ-Rain adopts a moisture-budget-guided process organization that separately models moisture storage, moisture transport and convergence, and surface evaporation, while incorporating precipitation persistence as an additional predictive signal. Moreover, to address the highly intermittent and zero-inflated characteristics of precipitation, MZ-Rain introduces an adaptive Tweedie-based multi-task learning framework that combines state-dependent quantitative precipitation modeling with auxiliary rainfall-occurrence supervision. In this way, MZ-Rain integrates physically motivated process organization with distribution-aware learning for station-level precipitation forecasting.
\subsection{Physics-Guided Meteorological Modeling}

Recent studies such as NeuralGCM \cite{kochkov2024neural} and WeatherGFT \cite{xu2024generalizing} have explored the integration of physical knowledge into neural meteorological models to improve interpretability, generalization, and physical consistency \cite{vermaclimode}. 
ClimODE \cite{vermaclimode} formulates climate and weather forecasting as a continuous-time neural ODE system based on advection dynamics. By modeling weather evolution as the transport of physical quantities through a learned neural flow, ClimODE introduces value-conserving dynamics and uncertainty estimation, showing the benefit of embedding conservation-inspired principles into deep forecasting models. 
PhyDL-NWP \cite{luo2025physics} further incorporates meteorological governing equations into deep learning by constructing differentiable PDE terms and introducing latent force parameterization to represent unresolved or unobserved physical processes. Its physics-guided loss improves both downscaling and forecasting by encouraging predictions to be consistent with parameterized meteorological dynamics. 
PINT \cite{park2025pint} introduces physics-guided neural time-series models for long-term climate forecasting. It uses the simple harmonic oscillator equation as a physical prior to capture periodic dynamics in 2m-temperature data and incorporates this constraint into recurrent architectures such as RNN, LSTM, and GRU. This demonstrates the effectiveness of embedding physical priors into temporal models.

These studies demonstrate that physical knowledge can improve neural forecasting through conservation principles, governing equations, or dynamics-based constraints. However, such physical knowledge is often incorporated indirectly through auxiliary losses, equation residuals, or dynamical regularization, while the learned representations themselves remain largely data-driven. Moreover, these approaches mainly target general weather or climate evolution rather than the process structure underlying station-level precipitation. In contrast, MZ-Rain explicitly uses the moisture budget to organize meteorological observations into distinct precipitation-related processes, introducing physical knowledge directly into the representation structure. This provides a precipitation-specific physical inductive bias for station-level forecasting.

\subsection{Zero-Inflated Modeling}

Statistical modeling of zero-inflated data has been extensively studied. Classical approaches, such as the Zero-Inflated Poisson (ZIP) model~\cite{lambert1992zero}, explicitly account for excessive zeros by combining a binary zero-generating process with a count regression process. Extensions to temporal settings, such as ZIPPAR~\cite{maiti2014modelling}, further model sparse count sequences with autoregressive structures. More recent studies have revisited zero-inflated Poisson regression and compared different zero-inflated models on modern datasets~\cite{beveridge2024comparison}, demonstrating their effectiveness in handling excess-zero observations. In addition, hybrid approaches that combine deep learning with zero-inflated statistical modeling have emerged in various domains. For example, Shi et al.~\cite{shi2021new} proposed a deep learning-based zero-inflated duration model for financial applications, showing the potential of integrating neural networks with zero-inflated formulations.

Unlike classical zero-inflated models that explicitly separate zero generation from positive-value regression, MZ-Rain handles precipitation intermittency through an adaptive Tweedie-based multi-task objective. The Tweedie formulation accommodates the mixed discrete-continuous nature of precipitation, while auxiliary rainfall-occurrence supervision strengthens discrimination between dry and rainy states. This allows zero inflation to be addressed within a unified forecasting framework rather than through a separate zero-generating mixture model. Together with the process-aware representation described above, MZ-Rain jointly addresses the physical heterogeneity and statistical intermittency of station-level precipitation.

\section{Preliminaries}
\label{sec:preliminary}
\subsection{Problem Statement}
We formulate station-level precipitation nowcasting as a multivariate-to-univariate time-series forecasting task. Given the historical sequence
$
\mathbf{X}=\{(\mathbf{x}_t,y_t)\}_{t=i}^{i+l-1}, \mathbf{x}_t\in\mathbb{R}^D,
$
where $\mathbf{x}_t$ denotes the meteorological variables at time $t$, $y_t$ is the observed precipitation, $D$ is the number of meteorological variables, and $l$ is the input length, the goal is to predict the future precipitation sequence
$
\hat{\mathbf{y}} = f(\mathbf{X}), \mathbf{y}=\{y_{i+l},y_{i+l+1},\dots,y_{i+l+H-1}\},
$
where $H$ is the prediction horizon.

\subsection{Moisture Budget Equation}

Precipitation formation is fundamentally governed by the column-integrated moisture budget, which describes the conservation of atmospheric water vapor. In this work, we express the moisture budget as
\begin{equation}
P_t = - \frac{\partial PWV_t}{\partial t} - \nabla \cdot \mathbf{Q}_t + E_t,
\label{eq:precip_from_budget}
\end{equation}
where $P_t$ denotes precipitation, $PWV_t$ is precipitable water vapor, $\frac{\partial PWV_t}{\partial t}$ represents the local tendency of column-integrated water vapor, $\mathbf{Q}_t$ denotes the vertically integrated horizontal moisture flux, $\nabla \cdot \mathbf{Q}_t$ represents horizontal moisture flux divergence, and $E_t$ denotes the surface moisture source associated with evaporation or evapotranspiration.

Equation~\eqref{eq:precip_from_budget} motivates a physically decomposed model design, where the storage, transport, and evaporation branches represent moisture storage change, horizontal moisture transport, and surface moisture supply, respectively. Specifically, the storage branch is associated with $-\frac{\partial PWV_t}{\partial t}$, the surface evaporation branch approximates $E_t$, and the transport branch provides a transport-related representation.

\subsection{Tweedie Distribution}
Precipitation exhibits a mixed discrete--continuous structure, with a point mass at zero and continuous, right-skewed positive values. To model this zero-inflated and heteroscedastic behavior, we adopt the Tweedie distribution:
$
Y\sim \mathrm{Tw}_p(\mu,\phi), 1<p<2,
$
where $\mu>0$ is the mean, $\phi>0$ is the dispersion parameter, and $p$ is the variance power. It satisfies
$
\mathbb{E}[Y]=\mu, \mathrm{Var}(Y)=\phi\mu^p.
$
When $1<p<2$, the Tweedie distribution admits a compound Poisson--Gamma form,
\begin{equation}
Y=\sum_{k=1}^{N} Z_k,\qquad N\sim\mathrm{Poisson}(\lambda),\qquad Z_k>0,
\end{equation}
which implies a nonzero probability mass at zero:
\begin{equation}
\mathbb{P}(Y=0)=e^{-\lambda}>0.
\end{equation}
These properties make it well suited for precipitation forecasting.
\subsection{Dataset and the study stations}
\label{sec:dataset}

\subsubsection{Dataset}
The RainfallBench~\cite{11424616} dataset is a multi-source
observational dataset designed for GNSS-based precipitation
nowcasting. It integrates ground-based atmospheric observations,
satellite-derived precipitation measurements, and ERA5 reanalysis
meteorological fields. The dataset spans the period from January
2018 to January 2024 and contains observations from six stations
located under different geographical and climatic conditions.

The dataset used in this study is constructed from three major data
sources. First, precipitable water vapor (PWV) is derived from GNSS
tropospheric products provided by the Nevada Geodetic Laboratory,
which processes observations from a global network of more than
19,000 GNSS stations. Second, precipitation is obtained from the
GPM IMERG Final Precipitation Product Version~07 generated by the
Global Precipitation Measurement mission. The IMERG precipitation
data are used both as historical precipitation inputs and as the
forecasting target. Third, ERA5 reanalysis products provide the
meteorological variables required to characterize the atmospheric
moisture conditions associated with precipitation, including surface
pressure, surface evaporation, and vertically integrated horizontal
water-vapor fluxes.

Following the moisture-budget-guided process organization described
in Section~\ref{sec:method}, the variables are reorganized into four
process-specific pathways. The moisture-storage pathway
(\textit{ms}) contains PWV $W_t$, its first-order temporal difference
$\Delta W_t$, and surface pressure $P_t^{\mathrm{sfc}}$. The
moisture-transport and convergence pathway (\textit{tr}) uses
vertically integrated moisture-flux convergence $C_t$. The
surface-evaporation pathway (\textit{ev}) uses surface evaporation
$E_t$, while historical precipitation $y_t$ is introduced through
the precipitation-persistence pathway (\textit{pr}).

For the moisture-transport and convergence pathway, we use the ERA5
vertically integrated eastward and northward water-vapor fluxes,
denoted by $Q^{x}$ and $Q^{y}$, respectively. Instead of directly
using near-surface wind speed as a proxy for moisture transport,
moisture-flux convergence is diagnosed from the spatially distributed
water-vapor flux field. The horizontal moisture-flux divergence is
calculated as

\begin{equation}
\nabla_h \cdot \mathbf{Q}
=
\frac{1}{R\cos\varphi}
\frac{\partial Q^{x}}{\partial \lambda}
+
\frac{1}{R\cos\varphi}
\frac{\partial \left(Q^{y}\cos\varphi\right)}
{\partial \varphi},
\label{eq:moisture_flux_divergence_dataset}
\end{equation}

where $R$ is the Earth's radius, $\lambda$ is longitude, and
$\varphi$ is latitude. Moisture-flux convergence is then defined as

\begin{equation}
C_t
=
-
\nabla_h \cdot \mathbf{Q}_t,
\label{eq:moisture_convergence_dataset}
\end{equation}

such that $C_t>0$ represents horizontal moisture convergence,
whereas $C_t<0$ represents moisture divergence. The divergence is
first calculated from the gridded ERA5 moisture-flux field, and the
resulting convergence field is subsequently interpolated to the
corresponding GNSS station location. In this way, although the
forecasting task is performed at the station level, the transport
pathway retains spatial information from the surrounding atmospheric
moisture-flux field.

Surface evaporation $E_t$ is used to characterize surface moisture
supply. In the processed dataset, evaporation is defined as positive
for upward moisture transfer from the surface to the atmosphere.
Therefore, positive $C_t$ and $E_t$ respectively indicate moisture
contributions associated with horizontal convergence and surface
evaporation.

To ensure consistency among heterogeneous data sources, all variables
are harmonized in both the temporal and spatial dimensions. The
original 5-minute GNSS observations and 30-minute IMERG precipitation
estimates are aggregated to an hourly temporal resolution. ERA5 reanalysis
variables are aligned to the same hourly timestamps. For station-level
meteorological variables, the gridded ERA5 reanalysis fields are
interpolated to the corresponding GNSS station locations using
bilinear interpolation. In contrast, IMERG precipitation is assigned
using the nearest-neighbor approach to avoid additional smoothing of
localized rainfall intensity. For moisture-flux convergence, the
spatial divergence is calculated before station-level interpolation.

The final input variables used by MZ-Rain are summarized in
Table~\ref{tab:variables}. All variables are provided at an hourly
temporal resolution.

\begin{table}[t]
\centering
\caption{Input variables used by MZ-Rain and their corresponding
data sources and process pathways.}
\label{tab:variables}
\small
\resizebox{\columnwidth}{!}{
\begin{tabular}{llll}
\toprule
\textbf{Variable} &
\textbf{Description} &
\textbf{Source} &
\textbf{Pathway} \\
\midrule

$W_t$
&
Precipitable water vapor
&
GNSS
&
ms
\\

$\Delta W_t$
&
Temporal difference of PWV
&
GNSS-derived
&
ms
\\

$P_t^{\mathrm{sfc}}$
&
Surface pressure
&
ERA5 Reanalysis
&
ms
\\

$C_t$
&
Moisture-flux convergence
&
ERA5-derived
&
tr
\\

$E_t$
&
Surface evaporation
&
ERA5 Reanalysis
&
ev
\\

$y_t$
&
Historical precipitation
&
IMERG
&
pr
\\

\bottomrule
\end{tabular}
}
\end{table}

\subsubsection{Split, Windowing, and Leakage Control}

We formulate station-level precipitation nowcasting as a multivariate-to-univariate sequence forecasting task. For each forecasting sample, the previous $L=24$ hourly observations are used as input to predict precipitation over the subsequent $H\in{2,4,6}$ hours. Separate models are trained for the 2-, 4-, and 6-h forecasting horizons under the same experimental protocol.

For each station, the observations are chronologically divided into training, validation, and test subsets using a ratio of $7{:}1{:}2$. All normalization statistics are estimated exclusively from the training subset and subsequently applied to the validation and test subsets. Sliding input--target windows are constructed independently within each split and are not allowed to cross split boundaries, thereby preventing information from future periods from entering model training. All compared methods use identical station-wise partitions and exactly the same input--target sample indices to ensure a fair comparison.

Model selection, including checkpoint selection and any validation-dependent thresholds, is performed exclusively using the validation subset. The test subset is reserved for final evaluation and is not used for model selection or hyperparameter tuning.

\subsubsection{Missing-Value Handling}

Missing values in continuous atmospheric variables, including PWV and ERA5 reanalysis-derived fields, are handled using linear interpolation for short temporal gaps to maintain the continuity of the input sequences. Precipitation targets are not forward-filled. Instead, forecasting windows containing missing target precipitation values are excluded from training and evaluation. This treatment avoids artificially extending observed rainfall states and prevents precipitation imputation from introducing an unintended advantage to the precipitation-persistence pathway.

\subsubsection{Study Stations}
\begin{figure}[!t]
    \centering
    \includegraphics[width=0.9\columnwidth]{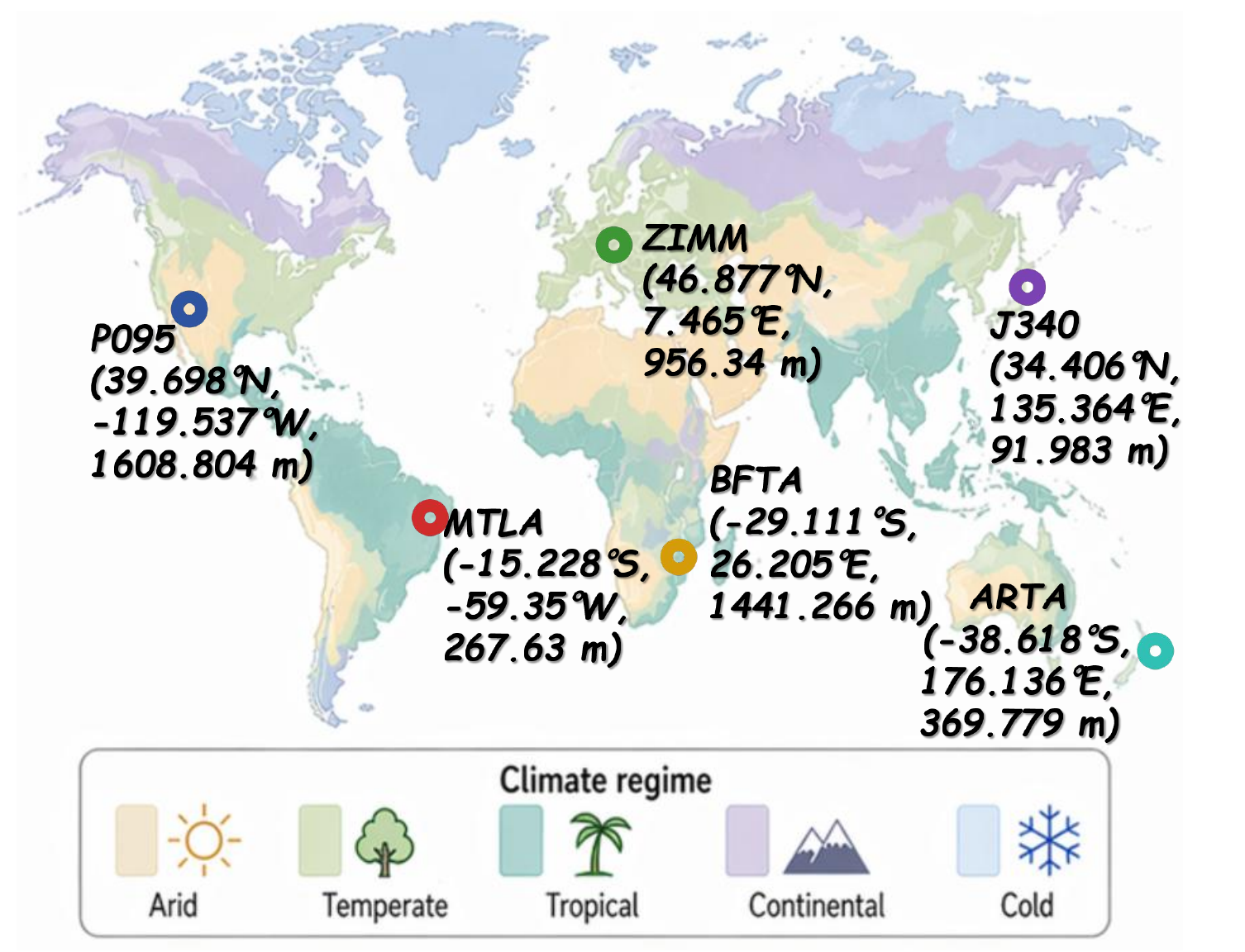}
    \caption{Geographical distribution of the six GNSS stations used in this study. The selected stations span diverse geographic locations and climatic conditions, providing broad spatial coverage for precipitation forecasting evaluation.}
    \label{fig:station-distribution}
\end{figure}

As shown in Fig.~\ref{fig:station-distribution}, the six GNSS stations used in our experiments are strategically selected to ensure broad geographic and climatic coverage.

{\color{black}\textbf{J340} (34.406°N, 135.364°E, 91.983 m)} – Located in the Kinki region of Japan, this station is situated at a low elevation. It is characterized by a humid subtropical climate (Cfa), with four distinct seasons, hot and rainy summers, mild winters, and relatively evenly distributed precipitation throughout the year. The region is also occasionally affected by typhoons in summer. 

{\color{black}\textbf{ZIMM} (46.877°N, 7.465°E, 956.34 m)} – Located in central Switzerland in the Alps, this mid-altitude station exhibits a temperate continental climate (Dfb), characterized by cold, snowy winters and warm, humid summers, with precipitation distributed throughout the year and frequent summer thunderstorms. 

{\color{black}\textbf{P095} (39.698°N, -119.537°W, 1608.804 m)} – Located in western Nevada, USA, this high-altitude station experiences a temperate desert climate (BWk), with arid conditions, large diurnal temperature variations, cold winters, hot summers, and low annual precipitation. 

{\color{black}\textbf{MTLA} (-15.228°S, -59.35°W, 267.63 m)} – Located in Mato Grosso, Brazil, this low-elevation station exhibits a tropical wet and dry climate (Aw/Am), with a pronounced wet season in summer (November–March) and a dry winter season. The region experiences high average annual temperatures. 

{\color{black}\textbf{ARTA} (-38.618°S, 176.136°E, 369.779 m)} – This station in the eastern North Island of New Zealand lies at a moderate elevation. The climate is temperate oceanic (Cfb), with mild and humid conditions year-round, evenly distributed precipitation, warm summers, and cool winters, strongly influenced by the surrounding ocean. 

{\color{black}\textbf{BFTA} (-29.111°S, 26.205°E, 1441.266 m)} – Situated in northern South Africa, this high-altitude station experiences a subtropical highland climate (Cwb), with warm and wet summers, cool and dry winters, and most precipitation occurring during the summer months.

\section{Method}
\label{sec:method}
\subsection{Overview of MZ-Rain}
MZ-Rain provides a unified framework that integrates physics-guided process representation with distribution-aware precipitation modeling, thereby accounting for both the physical and statistical structures of station-level precipitation.
Specifically, MZ-Rain consists of two key components. The first is a moisture-budget-guided multi-branch sLSTM encoder, which organizes
budget-related atmospheric information into pathways associated
with moisture storage, moisture transport and convergence, and
surface evaporation. 
The second is an adaptive Tweedie-based multi-task learning framework designed to address the zero-inflated nature of precipitation. It integrates quantitative precipitation modeling with an auxiliary rainfall-occurrence task, enabling complementary learning of precipitation amount and wet-dry discrimination within a unified forecasting framework.

\begin{figure*}[t!]
\centering
\includegraphics[width=0.85\textwidth]{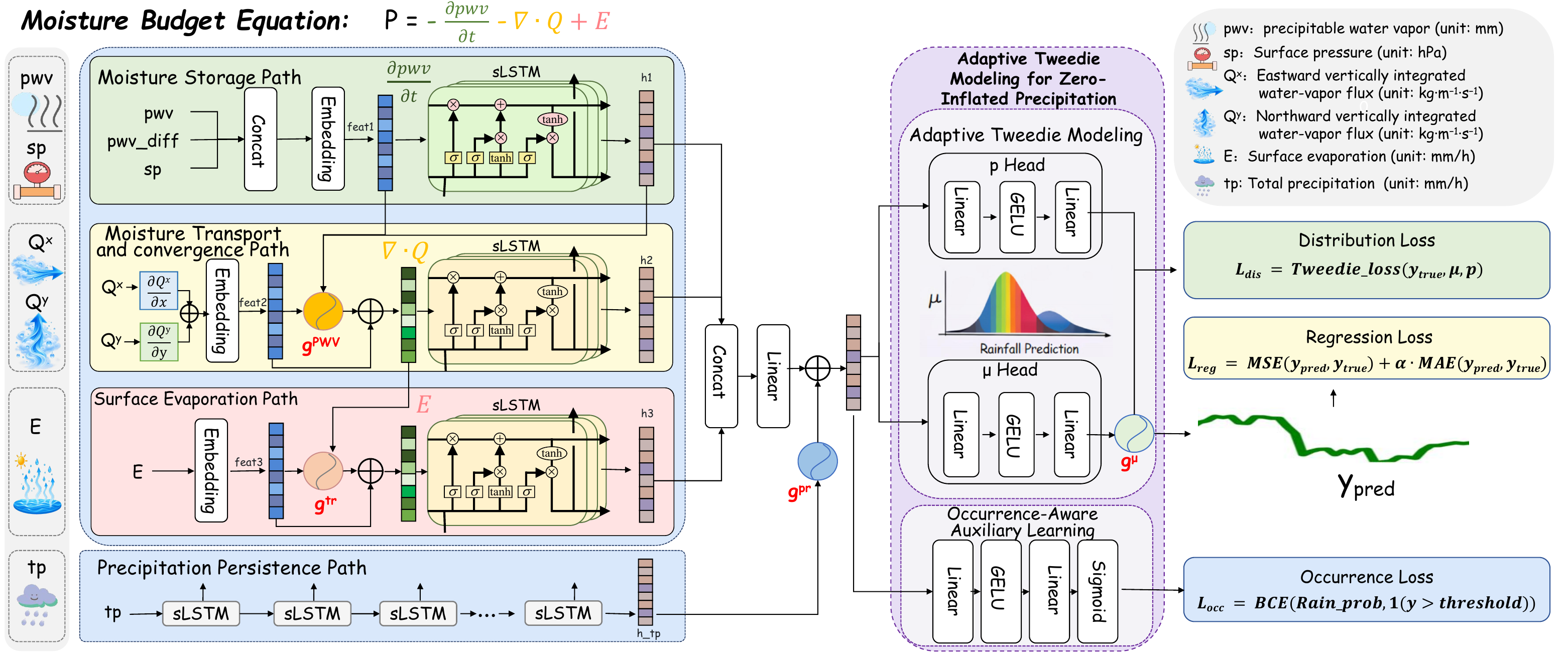}
\caption{Architecture of MZ-Rain. The left side shows four process-specific pathways for moisture storage, moisture transport and convergence, surface evaporation, and precipitation persistence. The right side shows adaptive Tweedie modeling for quantitative precipitation estimation together with auxiliary rainfall-occurrence supervision.
}
\label{fig:overall}
\end{figure*}

The overall architecture is illustrated in Figure~\ref{fig:overall}. The following sections introduce the
process-specific input organization, multi-branch temporal
encoding, adaptive Tweedie modeling, and the composite training
objective.
\subsection{Moisture-Budget-Guided Process Organization}

Following the moisture budget equation introduced in
Section~\ref{sec:preliminary}, precipitation evolution is closely
associated with three core atmospheric moisture processes: the
temporal variation of column moisture storage, the horizontal
transport and convergence of atmospheric moisture, and surface
moisture supply through evaporation. In addition, historical
precipitation provides an important persistence signal that is highly
informative for short-term nowcasting. Our goal is not to explicitly
solve the moisture budget equation or enforce exact budget closure,
but to use its process structure as a physical prior for organizing
temporal representation learning.

Based on this observation, we organize the input sequence into four
process-specific components:
\begin{equation}
\mathcal{X}
=
\left\{
\mathcal{X}^{\mathrm{ms}},
\mathcal{X}^{\mathrm{tr}},
\mathcal{X}^{\mathrm{ev}},
\mathcal{X}^{\mathrm{pr}}
\right\},
\end{equation}
where $\mathcal{X}^{\mathrm{ms}}$,
$\mathcal{X}^{\mathrm{tr}}$,
$\mathcal{X}^{\mathrm{ev}}$, and
$\mathcal{X}^{\mathrm{pr}}$ denote the moisture-storage,
moisture-transport and convergence, surface-evaporation, and
precipitation-persistence components, respectively.

For the moisture-storage component, we use precipitable water vapor
(PWV), its first-order temporal difference, and surface pressure:
\begin{equation}
\mathcal{X}^{\mathrm{ms}}_t
=
[W_t,\Delta W_t,P_t^{\mathrm{sfc}}],
\qquad
\Delta W_t=W_t-W_{t-1},
\end{equation}
where $W_t$ denotes PWV, $P_t^{\mathrm{sfc}}$ provides complementary information about the atmospheric column mass, helping contextualize the absolute and temporal variation of PWV under different background atmospheric states, and
$\Delta W_t$ characterizes the short-term variation of atmospheric
column moisture. We set $\Delta W_0=0$.

For the moisture-transport and convergence component, we use
moisture-flux convergence $C_t$ to characterize the net horizontal
moisture accumulation associated with the vertically integrated
moisture transport. As described in the dataset construction,
$C_t$ is diagnosed from the vertically integrated horizontal
moisture-flux field as
\begin{equation}
C_t
=
-\nabla_h\cdot\mathbf{Q}_t,
\end{equation}
where $\mathbf{Q}_t$ denotes the vertically integrated horizontal
moisture-flux vector. With this sign convention, $C_t>0$ represents
horizontal moisture convergence, whereas $C_t<0$ represents moisture
divergence. The corresponding pathway input is therefore defined as
\begin{equation}
\mathcal{X}^{\mathrm{tr}}_t
=
[C_t].
\end{equation}

Although the forecasting target is defined at an individual station,
$C_t$ is diagnosed from the spatially distributed vertically
integrated moisture-flux field surrounding the station before being
provided to the temporal forecasting model. Therefore, the
moisture-transport pathway directly incorporates spatial moisture
convergence information rather than inferring it from a local scalar
wind observation.

For the surface-moisture-supply component, we directly use surface
evaporation:
\begin{equation}
\mathcal{X}^{\mathrm{ev}}_t=[E_t],
\end{equation}
where $E_t$ represents the upward surface moisture flux associated
with evaporation. This provides a direct measure of surface moisture
supply to the atmospheric column.

Finally, recent precipitation is introduced as an additional
predictive-memory signal:
\begin{equation}
\mathcal{X}^{\mathrm{pr}}_t=[y_t],
\end{equation}
where $y_t$ denotes historical precipitation. Unlike the first three
components, the precipitation-persistence component is not an
independent term of the moisture budget, but is included to capture
the strong short-term continuity and autocorrelation of precipitation
events.

Accordingly, the first three components provide process-specific
atmospheric information motivated by the moisture budget, while the
precipitation-persistence component complements them with recent
rainfall memory. These components are subsequently modeled by
dedicated temporal branches to capture their heterogeneous temporal
dynamics.

\subsection{Process-Specific Multi-Branch Temporal Encoding}
Based on the moisture-budget-guided process organization, we design a multi-branch sLSTM architecture to model the temporal evolution of three process-specific atmospheric signals associated with moisture storage, moisture transport and convergence, and surface evaporation.
Rather than encoding all meteorological variables with a single shared temporal backbone, MZ-Rain assigns a dedicated sLSTM branch to each process pathway, so that their heterogeneous dynamics can be modeled in a process-specific manner.

\paragraph{Moisture Storage Pathway}
The first branch models the temporal evolution of atmospheric moisture storage through precipitable water vapor and its short-term tendency. We therefore construct the storage pathway from precipitable water vapor, its first-order difference, and surface pressure as an atmospheric background variable:
$
\mathcal{X}^{\mathrm{ms}}_t = [W_t, \Delta W_t, P_t^{\mathrm{sfc}}], 
\Delta W_t = W_t - W_{t-1},
$
where $W_t$ denotes precipitable water vapor and $P_t^{\mathrm{sfc}}$ denotes surface pressure. After a learnable embedding transformation $\Phi_{\mathrm{ms}}(\cdot)$, the corresponding latent representation is obtained by
\begin{equation}
\mathbf{h}^{\mathrm{ms}} = \mathrm{Enc}_{\mathrm{ms}}\!\left(\Phi_{\mathrm{ms}}(\mathcal{X}^{\mathrm{ms}})\right),
\end{equation}
where $\mathrm{Enc}_{\mathrm{ms}}(\cdot)$ denotes the sLSTM encoder for the moisture storage pathway.

\paragraph{Moisture Transport and Convergence Pathway}
The second branch models moisture transport and convergence, which characterize the horizontal redistribution and local accumulation or depletion of atmospheric water vapor. Importantly, the effect of moisture convergence depends on the prevailing moisture background, since similar convergence states may have different implications under dry and moist atmospheric conditions.
To reflect this dependency, we introduce a \emph{PWV-guided gating mechanism} that allows the storage state to modulate the moisture transport and convergence pathway.

We first compute a gate from the storage representation,
\begin{equation}
\mathbf{g}^{\mathrm{pwv}} = \sigma\!\left(\mathbf{W}_{\mathrm{pwv}}\mathbf{h}^{\mathrm{ms}}\right),
\end{equation}
and use it to inject moisture-background information into the transport branch:
\begin{equation}
\tilde{\mathcal{X}}^{\mathrm{tr}} =
\Phi_{\mathrm{tr}}(\mathcal{X}^{\mathrm{tr}})
+
\mathbf{g}^{\mathrm{pwv}} \odot \Phi_{\mathrm{ms}}(\mathcal{X}^{\mathrm{ms}}).
\end{equation}
The transport representation is then given by
\begin{equation}
\mathbf{h}^{\mathrm{tr}} = \mathrm{Enc}_{\mathrm{tr}}\!\left(\tilde{\mathcal{X}}^{\mathrm{tr}}\right),
\end{equation}
where $\mathrm{Enc}_{\mathrm{tr}}(\cdot)$ denotes the transport sLSTM encoder. This \textbf{PWV gate} explicitly encodes the dependence of transport dynamics on the moisture storage state, making the transport pathway conditional on the prevailing atmospheric moisture background.

\paragraph{Surface Evaporation Pathway}
The third branch models surface moisture supply associated with evaporation. In the moisture budget, surface evaporation represents an upward moisture flux from the surface to the atmosphere and provides an additional source of atmospheric water vapor.
Since the contribution of surface moisture supply may depend on the prevailing moisture-transport conditions, we introduce a \emph{transport-guided gating mechanism}.

After embedding the surface-evaporation input with $\Phi_{\mathrm{ev}}(\cdot)$, we compute a transport-dependent gate
\begin{equation}
\mathbf{g}^{\mathrm{tr}} = \sigma\!\left(\mathbf{W}_{\mathrm{tr}}\mathbf{h}^{\mathrm{tr}}\right),
\end{equation}
and modulate the surface-evaporation pathway as
\begin{equation}
\tilde{\mathcal{X}}^{\mathrm{ev}} =
\Phi_{\mathrm{ev}}(\mathcal{X}^{\mathrm{ev}})
+
\mathbf{g}^{\mathrm{tr}} \odot \Phi_{\mathrm{ev}}(\mathcal{X}^{\mathrm{ev}}).
\end{equation}
The surface-evaporation representation is then obtained by
\begin{equation}
\mathbf{h}^{\mathrm{ev}} = \mathrm{Enc}_{\mathrm{ev}}\!\left(\tilde{\mathcal{X}}^{\mathrm{ev}}\right).
\end{equation}
The resulting \textbf{transport gate} enables the model to adapt the surface-evaporation representation according to the prevailing moisture-transport regime, thereby allowing interactions between horizontal moisture transport and surface moisture supply to be modeled adaptively.

\paragraph{Fusion of Moisture-Budget Pathways}
The process-specific representations are concatenated and linearly projected into a unified moisture-budget-guided representation:
\begin{equation}
\mathbf{h}^{\mathrm{mb}} =
\Psi\!\left(
[\mathbf{h}^{\mathrm{ms}};\mathbf{h}^{\mathrm{tr}};\mathbf{h}^{\mathrm{ev}}]
\right),
\end{equation}
where $\Psi(\cdot)$ denotes a linear projection.

\paragraph{Precipitation persistence}
Beyond the three moisture-budget-related pathways, short-term precipitation forecasting also strongly depends on rainfall persistence, since recent precipitation provides informative cues about event continuity and local temporal autocorrelation. As this information is not explicitly characterized by the moisture budget equation, we model it with an additional precipitation persistence branch.

Let $\mathcal{X}^{\mathrm{pr}}_t = [y_t]$ denote the historical precipitation input. Its latent representation is computed by a dedicated sLSTM encoder,
\begin{equation}
\mathbf{h}^{\mathrm{pr}} = \mathrm{Enc}_{\mathrm{pr}}\!\left(\Phi_{\mathrm{pr}}(\mathcal{X}^{\mathrm{pr}})\right).
\end{equation}
To adaptively control the contribution of rainfall memory, we further introduce a \textbf{precipitation-memory gate}
\begin{equation}
\mathbf{g}^{\mathrm{pr}} = \sigma\!\left(\mathbf{W}_{\mathrm{pr}}\mathbf{h}^{\mathrm{pr}}\right),
\end{equation}
which determines how strongly the persistence representation should influence the final prediction. The final latent representation is then obtained by
\begin{equation}
\mathbf{h} = \mathbf{h}^{\mathrm{mb}} + \mathbf{g}^{\mathrm{pr}} \odot \mathbf{h}^{\mathrm{pr}},
\end{equation}
where $\mathbf{h}^{\mathrm{mb}}$ denotes the fused moisture-budget-aware representation. 
In this way, the model combines moisture-budget-guided process representations with adaptively weighted rainfall-memory information, yielding a forecasting backbone that captures both atmospheric process dynamics and short-term precipitation persistence.

\subsection{Adaptive Tweedie Modeling for Zero-Inflated Precipitation}

While the moisture-budget-guided encoder captures process-specific
atmospheric dynamics, precipitation forecasting remains challenging
because rainfall observations are highly intermittent and exhibit a
mixed-support, strongly skewed, and heteroscedastic distribution.
To address this structure, MZ-Rain adopts an adaptive Tweedie-based
multi-task learning framework that combines quantitative
precipitation modeling with auxiliary rainfall-occurrence learning.

\paragraph{Adaptive Tweedie Modeling}
The Tweedie family provides a natural distributional inductive bias
for precipitation because, for $1<p<2$, it accommodates nonnegative
responses with a point mass at zero and continuous positive outcomes.

Given the fused forecasting representation $\mathbf h$, we first
construct a positive rainfall-amount candidate:
\begin{equation}
\tilde{\boldsymbol{\mu}}
=
\operatorname{softplus}
\left(
f_{\mu}(\mathbf h)
\right)
+\epsilon,
\label{eq:raw_mean}
\end{equation}
where $\tilde{\boldsymbol{\mu}}\in\mathbb{R}^{H}$ denotes the
ungated positive rainfall-amount candidate and $\epsilon>0$ is a
small constant for numerical stability.

To adapt the quantitative rainfall estimate to the prevailing
forecasting state, we introduce a data-dependent modulation
coefficient:
\begin{equation}
\mathbf g^{\mu}
=
\sigma
\left(
{W}_{\mathrm{\mu}}\mathbf h
\right),
\label{eq:mean_gate}
\end{equation}
where $ g^{\mu}\in(0,1)^H$ provides a horizon-wise
modulation of the rainfall-amount candidate.

The final nonnegative mean prediction is then obtained through
multiplicative modulation:
\begin{equation}
\hat{\mu}
=
\mathbf g^{\mu}
\odot
\tilde{\boldsymbol{\mu}},
\label{eq:adaptive_mean}
\end{equation}
where $\boldsymbol{\mu}\in\mathbb{R}^{H}_{+}$ is the final
quantitative precipitation prediction used by the point-wise and
Tweedie-based objectives.

The modulation coefficient provides a state-dependent mechanism for
adjusting the contribution of the raw rainfall-amount candidate.
Rather than introducing a separate probabilistic interpretation,
$\mathbf g^{\mu}$ serves as an architectural calibration factor whose
effect is realized through the final mean $\boldsymbol{\mu}$.

The adaptive Tweedie power index is predicted as
\begin{equation}
\hat{p}
=
1+0.999\cdot
\sigma
\left(
f_p(\mathbf h)
\right),
\label{eq:power}
\end{equation}
so that $1<p_\tau<2$ for every forecasting step $\tau$.

For $1<p_\tau<2$, the resulting Tweedie-style objective captures
the nonnegative, mixed-support, and mean-dependent variability
characteristic of precipitation.
The quantitative precipitation forecast at each future step is
defined directly by the final mean:
\begin{equation}
\hat y_\tau=\hat{\mu}_\tau,
\qquad
\tau=1,\ldots,H.
\label{eq:final_prediction}
\end{equation}
Thus, the quantity optimized by the quantitative precipitation
objectives and used for inference is the same final mean prediction.

\paragraph{Occurrence-Aware Auxiliary Learning}
Although the Tweedie family inherently accommodates zero-valued observations, explicitly modeling rainfall occurrence remains beneficial for station-level precipitation nowcasting, since distinguishing dry and wet conditions is itself a critical prediction task. We therefore introduce an auxiliary occurrence branch that predicts the probability of rainfall at each future step:
\begin{equation}
\hat{q}
=
\sigma
\left(
f_{\mathrm{occ}}(\mathbf{h})
\right).
\end{equation}
where $\hat q_\tau\in(0,1)$ denotes the predicted probability of precipitation occurrence at forecasting step $\tau$. 
The occurrence prediction therefore provides auxiliary wet-dry supervision without directly entering the quantitative precipitation forecast.

Together, the quantitative precipitation and occurrence tasks provide
complementary supervision: the former focuses on accurate rainfall
amount estimation, while the latter emphasizes discrimination
between dry and wet conditions.

\subsection{Training Objective}

We jointly optimize wet-dry occurrence discrimination,
Tweedie-inspired quantitative precipitation modeling, and point-wise
regression. Specifically, letting $r_{\tau}=\mathbb{I}(y_{\tau}>\delta)$ denote the binary rainfall occurrence label, the overall objective is defined as
\begin{equation}
\mathcal{L}
=
\mathcal{L}_{\mathrm{MSE}}
+\beta \mathcal{L}_{\mathrm{MAE}}
+\alpha \mathcal{L}_{\mathrm{Tw}}
+\alpha \mathcal{L}_{\mathrm{BCE}},
\end{equation}
where
\begin{equation}
\mathcal{L}_{\mathrm{BCE}}
=
\frac{1}{H}\sum_{\tau=1}^{H}\mathrm{BCE}(\hat{q}_{\tau}, r_{\tau}),
\quad
\mathcal{L}_{\mathrm{Tw}}
=
\frac{1}{H}\sum_{\tau=1}^{H}\ell_{\mathrm{Tw}}(y_{\tau}; \hat{\mu}_{\tau}, \hat{p}_{\tau}),
\end{equation}
\begin{equation}
\mathcal{L}_{\mathrm{MSE}}
=
\frac{1}{H}\sum_{\tau=1}^{H}(\hat{y}_{\tau}-y_{\tau})^2,
\quad
\mathcal{L}_{\mathrm{MAE}}
=
\frac{1}{H}\sum_{\tau=1}^{H}|\hat{y}_{\tau}-y_{\tau}|.
\end{equation}
\begin{equation}
\ell_{\mathrm{Tw}}
\left(
y_{\tau};
\hat{\mu}_{\tau},
\hat{p}_{\tau}
\right)
=
-
\frac{
y_{\tau}
(\hat{\mu}_{\tau}+\epsilon)^{1-\hat{p}_{\tau}}
}
{
1-\hat{p}_{\tau}+\epsilon
}
+
\frac{
(\hat{\mu}_{\tau}+\epsilon)^{2-\hat{p}_{\tau}}
}
{
2-\hat{p}_{\tau}+\epsilon
},
\end{equation}
Here, $\delta = 0.1~\mathrm{mm/h}$ defines the dry-wet threshold, $\epsilon = 10^{-6}$ ensures numerical stability, and $\alpha=0.1$ and $\beta=0.3$ are loss balancing coefficients.

\section{Experiments}
\label{sec:experiments}

Our experiments are organized around two evidence chains that mirror the
two technical contributions of MZ-Rain.
The first chain (Section~\ref{sec:exp-process}) asks whether the
moisture-budget variables carry useful process information, whether
removing them degrades forecasts, and whether the physically motivated
process organization outperforms generic multi-branch alternatives.
The second chain (Section~\ref{sec:exp-tweedie}) asks whether the
Tweedie objective, the adaptive mean modulation, and the occurrence
auxiliary task each contribute.
Section~\ref{sec:exp-overall} reports the overall forecasting performance, while Section~\ref{sec:exp-heavy} provides a dedicated evaluation under heavy-rain conditions. Finally, Section~\ref{app:qualitative} presents a qualitative analysis of representative precipitation cases to illustrate the forecasting behavior of MZ-Rain under different rainfall regimes.

\subsection{Experimental Setup}
\label{sec:exp-setup}

\subsubsection{Dataset and Forecasting Protocol}
\label{sec:exp-dataset}

We construct hourly precipitation forecasting datasets at six GNSS
stations spanning six years (January 2018 to January 2024) under
diverse geographical and climatic conditions.
Each sample uses a 24-hour history to forecast precipitation over
horizons of 2, 4, and 6 hours. Further details on the study sites, data sources, preprocessing procedures, and dataset construction are provided in Section \ref{sec:dataset} .

\subsubsection{Compared Methods}
\label{sec:exp-baselines}

We select baseline models from three categories for a comprehensive comparison.

% \textbf{(i) General time-series forecasting models.}
% We include sLSTM \cite{beck2024xlstm}, TimeFilter \cite{hu2025timefilter},
% TimeKAN \cite{huang2025timekan}, xPatch \cite{stitsyuk2025xpatch},
% and FilterTS \cite{wang2025filterts}, covering RNN-based, GNN-based,
% KAN-based, CNN-based, and MLP-based architectures, respectively.

% \textbf{(ii) Precipitation-specific forecasting model.}
% We include BFPF \cite{11424616}, an Transformer-based
% \cite{zhou2021informer} precipitation forecasting model designed
% to handle sparse and zero-inflated rainfall sequences.

% \textbf{(iii) Statistical model for zero-inflated data.}
% We include the Zero-Inflated Poisson (ZIP) model
% \cite{lambert1992zero}, a classical statistical model for
% zero-inflated data, to evaluate the effectiveness of explicitly
% modeling excessive zero-precipitation events.

% For a fair comparison, all deep learning baselines are implemented using
% the optimal configurations provided in their official repositories.
% We select baseline models from three categories for a comprehensive
% comparison.
\textbf{(i) General time-series forecasting models.}
We include sLSTM \cite{beck2024xlstm}, TimeFilter \cite{hu2025timefilter},
TimeKAN \cite{huang2025timekan}, xPatch \cite{stitsyuk2025xpatch},
and FilterTS \cite{wang2025filterts}, covering RNN-based, GNN-based,
KAN-based, CNN-based, and MLP-based architectures, respectively.

\textbf{(ii) Precipitation-specific forecasting models.}
We include BFPF \cite{11424616}, a Transformer-based
\cite{zhou2021informer} precipitation forecasting model designed to
handle sparse and zero-inflated rainfall sequences, and ZIDF
\cite{gao2025noise}, a recent diffusion-driven framework that couples a
non-stationary Transformer predictor with a denoising diffusion module,
specifically designed for zero-inflated precipitation forecasting.

\textbf{(iii) Statistical models for zero-inflated data.}
We include the Zero-Inflated Poisson (ZIP) model \cite{lambert1992zero},
a classical statistical model for zero-inflated data, and the
Hurdle-Gamma model \cite{mullahy1986specification}, which decomposes precipitation into a Bernoulli
occurrence stage and a Gamma intensity stage. Together they assess
whether explicitly modeling excessive zero-precipitation events
transfers to hourly rainfall nowcasting.

For a fair comparison, all deep learning baselines are implemented
using the optimal configurations provided in their official
repositories.

\subsubsection{Evaluation Metrics}

We evaluate forecasting performance under multiple prediction horizons. Let the set of output lengths be
$
\mathcal{L}_{\mathrm{out}} = \{2,4,6\},
$
where each $L_{\mathrm{out}} \in \mathcal{L}_{\mathrm{out}}$ corresponds to a forecasting horizon of 2, 4, and 6 hours, respectively. For a given horizon $L_{\mathrm{out}}$, let $\hat{\mathbf{y}}_{1:L_{\mathrm{out}}}$ and $\mathbf{y}_{1:L_{\mathrm{out}}}$ denote the predicted and ground-truth precipitation sequences.

\paragraph{Regression metrics}
To assess overall pointwise forecasting accuracy, we report the mean squared error (MSE) and mean absolute error (MAE), defined as
\begin{equation}
\mathrm{MSE}
=
\frac{1}{L_{\mathrm{out}}}
\sum_{t=1}^{L_{\mathrm{out}}}
\left(\hat{y}_t - y_t\right)^2,
\end{equation}
\begin{equation}
\mathrm{MAE}
=
\frac{1}{L_{\mathrm{out}}}
\sum_{t=1}^{L_{\mathrm{out}}}
\left|\hat{y}_t - y_t\right|.
\end{equation}

\paragraph{Event-based metrics}
Since precipitation forecasting also requires accurate discrimination between wet and dry conditions, we further adopt the Critical Success Index (CSI) and False Alarm Ratio (FAR). Let $\delta$ denote the threshold used to define a rainfall event, and define the binary event indicators as
$
\hat{r}_t = \mathbb{I}(\hat{y}_t > \delta),
r_t = \mathbb{I}(y_t > \delta).
$
Based on the corresponding contingency counts, namely hits ($H$), false alarms ($F$), and misses ($M$), CSI and FAR are computed as
\begin{equation}
\mathrm{CSI}
=
\frac{H}{H+F+M},
\end{equation}
\begin{equation}
\mathrm{FAR}
=
\frac{F}{H+F}.
\end{equation}
A higher CSI and a lower FAR indicate better event-level forecasting skill.

\paragraph{Extreme Precipitation Evaluation}

To further assess model performance on high-impact rainfall events, we conduct a dedicated evaluation on extreme precipitation cases. Following the T/CMSA 0013-2019 standard\footnote{\url{http://www.chinamsa.org/uploads/file/20191106142922_61962.pdf}}, we consider two severity levels:
\begin{equation}
\mathcal{E}^{(1)} = \{t \mid 10 \le y_t < 50\},
\qquad
\mathcal{E}^{(2)} = \{t \mid y_t \ge 50\},
\end{equation}
which correspond to heavy precipitation and very extreme precipitation, respectively.

For each subset $\mathcal{E}^{(k)}$, $k \in \{1,2\}$, we compute range-specific squared and absolute errors:
\begin{equation}
\mathrm{MSE}^{(k)}_{\mathrm{ext}}
=
\frac{1}{|\mathcal{E}^{(k)}|}
\sum_{t \in \mathcal{E}^{(k)}}
\left(\hat{y}_t - y_t\right)^2,
\end{equation}
\begin{equation}
\mathrm{MAE}^{(k)}_{\mathrm{ext}}
=
\frac{1}{|\mathcal{E}^{(k)}|}
\sum_{t \in \mathcal{E}^{(k)}}
\left|\hat{y}_t - y_t\right|.
\end{equation}
Lower values of $\mathrm{MSE}^{(k)}_{\mathrm{ext}}$ and $\mathrm{MAE}^{(k)}_{\mathrm{ext}}$ indicate better predictive accuracy on extreme precipitation events. This setting allows us to separately examine model behavior under moderately extreme and highly extreme rainfall regimes. 

\subsubsection{Implementation and Statistical Protocol}
\label{sec:exp-implementation}

All models are trained on a single NVIDIA H100 (80\,GB) with the Adam
optimizer; the architecture follows Section\ref{sec:method} with a 24-step input,
d\_model $=128$, embedding dimension $64$, and three sLSTM blocks per branch.
Unless stated otherwise, every reported number is the mean over five random seeds. 

\subsection{Overall Forecasting Performance}
\label{sec:exp-overall}

\begin{figure*}[t]
\centering
\includegraphics[width=\linewidth]{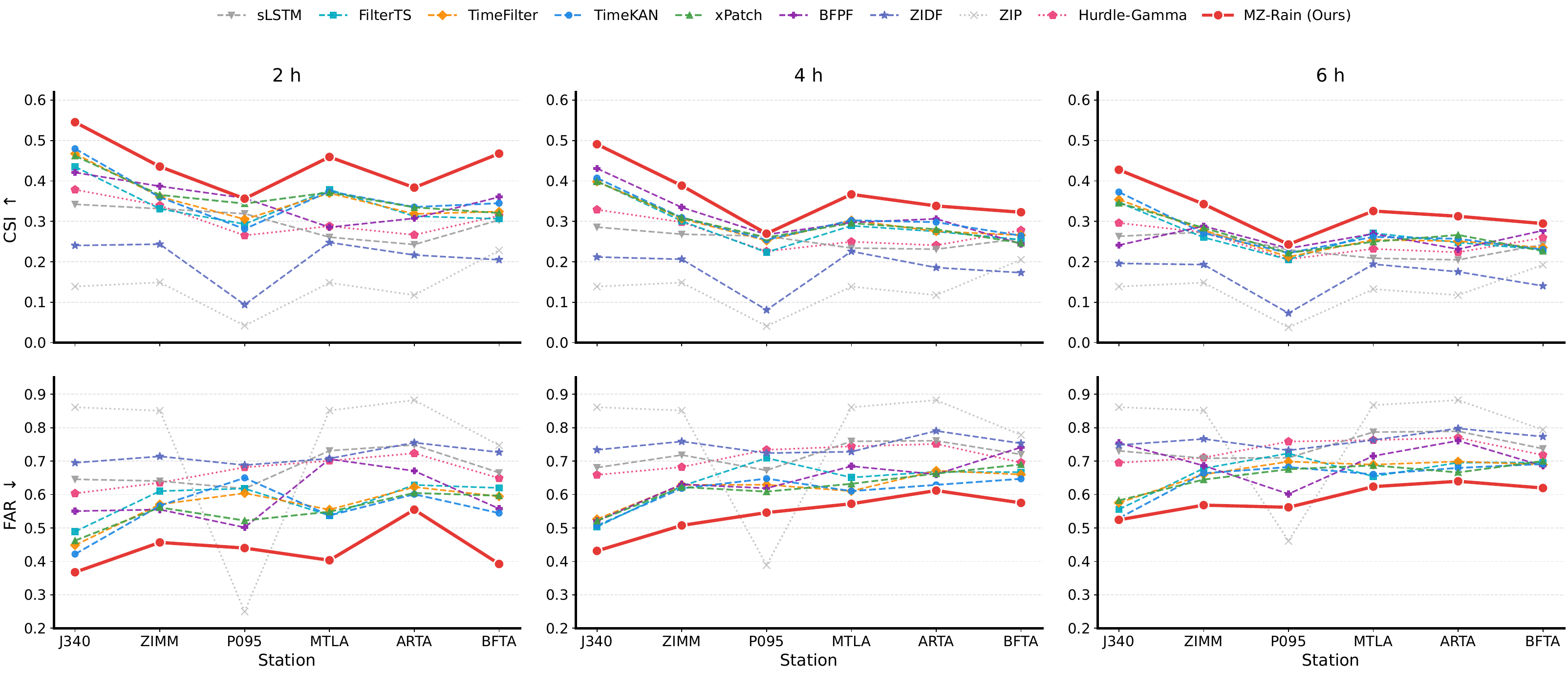}
\caption{Per-station detection performance at 2/4/6\,h horizons.
Each panel shows all ten models across the six GNSS stations
(J340, ZIMM, P095, MTLA, ARTA, BFTA); the red solid line is
\textbf{MZ-Rain (Ours)}.
\textbf{Top:} CSI ($\uparrow$, higher is better). \textbf{Bottom:}
FAR ($\downarrow$, lower is better). MZ-Rain attains the highest CSI
at every station and every horizon and the lowest FAR in nearly all
cases; at P095, the lower FAR of ZIP is accompanied by an extremely low CSI, indicating severe under-detection of precipitation events(cf.\ Figure~\ref{fig:station-lines-detection}  and Figure~\ref{fig:station-lines-intensity}).}
\label{fig:station-lines-detection}
\end{figure*}

\begin{figure*}[t]
\centering
\includegraphics[width=\linewidth]{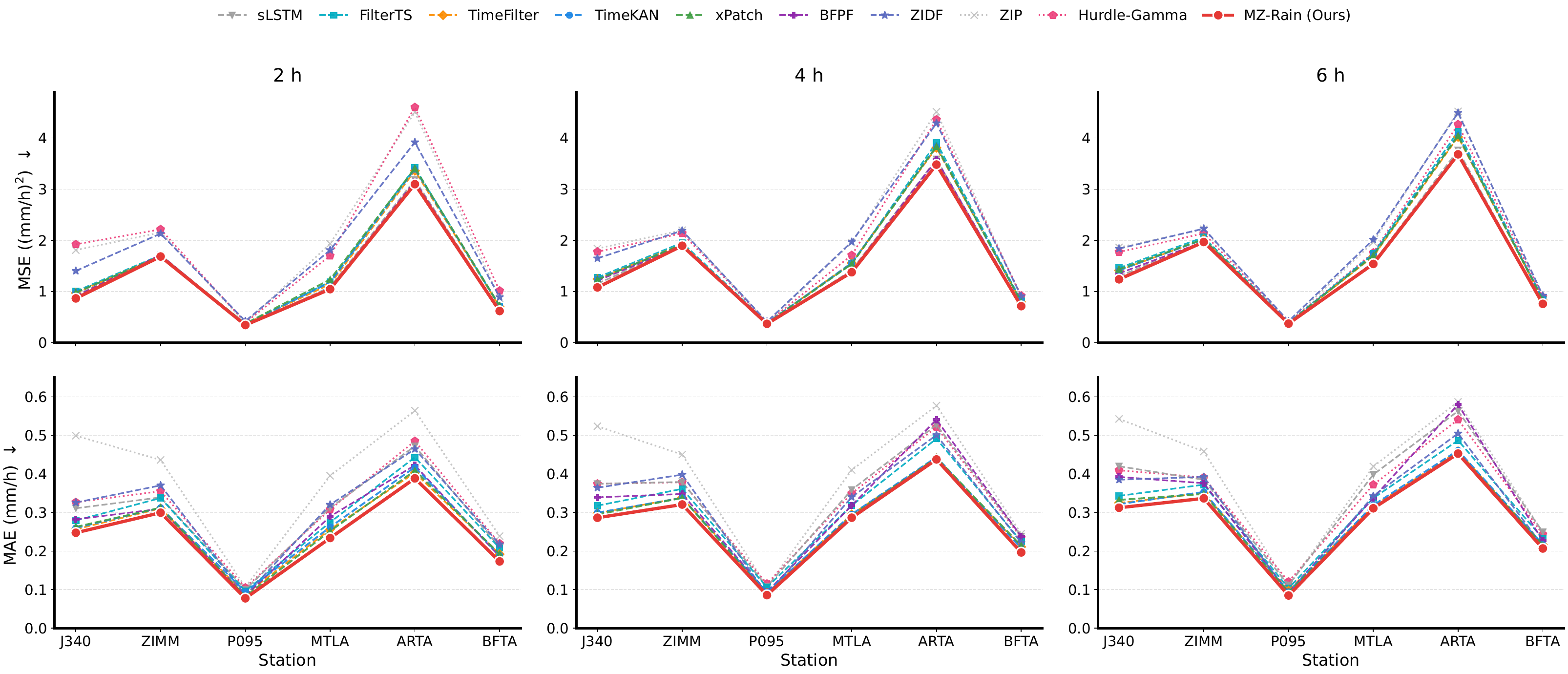}
\caption{Per-station intensity error at 2/4/6\,h horizons.
\textbf{Top:} MSE ((mm/h)$^2$, $\downarrow$). \textbf{Bottom:} MAE
(mm/h, $\downarrow$). MZ-Rain (red) achieves the lowest MAE at all
six stations and all horizons, and the lowest MSE in nearly all
cases.}
\label{fig:station-lines-intensity}
\end{figure*}

We first evaluate the overall forecasting performance of MZ-Rain against
all competing methods at prediction horizons of 2, 4, and 6\,h.
Table~\ref{tab:macro-results} summarizes the macro-averaged results over
the six stations, while Figure~\ref{fig:station-lines-detection}  and Figure~\ref{fig:station-lines-intensity} provide the
station-wise breakdown.

\begin{table*}
[t]
\centering
\caption{Cross-station macro-average performance at 2/4/6\,h horizons.
CSI$\uparrow$, FAR$\downarrow$, MSE$\downarrow$, and MAE$\downarrow$.
Avg.\ Rank denotes the mean rank over all twelve horizon--metric
combinations, where a lower value is better.}
\label{tab:macro-results}
\small
\setlength{\tabcolsep}{4.2pt}
\resizebox{\linewidth}{!}{
\begin{tabular}{l|cccc|cccc|cccc|c}
\toprule
\multirow{2}{*}{\textbf{Model}} &
\multicolumn{4}{c|}{\textbf{2\,h}} &
\multicolumn{4}{c|}{\textbf{4\,h}} &
\multicolumn{4}{c|}{\textbf{6\,h}} &
\multirow{2}{*}{\textbf{Avg.\ Rank}} \\
& MAE $\downarrow$ & MSE $\downarrow$ & CSI $\uparrow$& FAR$\downarrow$ &
  MAE$\downarrow$ & MSE$\downarrow$ & CSI$\uparrow$ & FAR$\downarrow$ &
  MAE$\downarrow$ & MSE$\downarrow$ & CSI$\uparrow$ & FAR$\downarrow$ & \\
\midrule
sLSTM
& 0.2909 & 1.3102 & 0.2999 & 0.6748
& 0.3311 & 1.5199 & 0.2567 & 0.7186
& 0.3547 & 1.6203 & 0.2370 & 0.7438 & 6.67 \\

FilterTS
& 0.2732 & 1.3985 & 0.3428 & 0.5839
& 0.3033 & 1.6486 & 0.2904 & 0.6373
& 0.3127 & 1.7699 & 0.2603 & 0.6671 & 5.58 \\

TimeFilter
& 0.2510 & 1.3726 & 0.3577 & 0.5658
& 0.2794 & 1.6178 & 0.3002 & 0.6210
& 0.2941 & 1.7262 & 0.2641 & 0.6686 & 3.75 \\

TimeKAN
& 0.2559 & 1.3848 & 0.3625 & 0.5535
& 0.2796 & 1.6200 & 0.3060 & 0.6098
& 0.2945 & 1.7386 & 0.2690 & 0.6511 & 3.67 \\

xPatch
& 0.2517 & 1.4010 & 0.3664 & 0.5493
& 0.2780 & 1.6162 & 0.2984 & 0.6228
& 0.2928 & 1.7369 & 0.2657 & 0.6593 & 3.50 \\

\midrule
BFPF
& 0.2624 & 1.3001 & 0.3531 & 0.5903
& 0.3131 & 1.5271 & 0.3135 & 0.6428
& 0.3344 & 1.6224 & 0.2567 & 0.7008 & 4.75 \\
ZIDF         & 0.2974 & 1.7623 & 0.2079 & 0.7144 & 0.3187 & 1.9028 & 0.1806 & 0.7480 & 0.3212 & 1.9858 & 0.1620 & 0.7635 & 8.50 \\

\midrule
ZIP
& 0.3734 & 1.9429 & 0.1374 & 0.7405
& 0.3858 & 1.9618 & 0.1316 & 0.7705
& 0.3943 & 1.9724 & 0.1279 & 0.7862 & 9.83 \\
Hurdle-Gamma & 0.3000 & 1.9742 & 0.3086 & 0.6656 & 0.3290 & 1.8829 & 0.2702 & 0.7110 & 0.3463 & 1.8699 & 0.2483 & 0.7358 & 7.75 \\
\midrule
\textbf{MZ-Rain}
& \textbf{0.2367} & \textbf{1.2762} & \textbf{0.4412} & \textbf{0.4358}
& \textbf{0.2690} & \textbf{1.4855} & \textbf{0.3627} & \textbf{0.5407}
& \textbf{0.2841} & \textbf{1.5929} & \textbf{0.3243} & \textbf{0.5895}
& \textbf{1.00} \\
\bottomrule
\end{tabular}}
\end{table*}

\subsubsection{Cross-Station Average Performance}
\label{sec:exp-macro}

A clear advantage of MZ-Rain is observed across both precipitation-amount
and precipitation-event metrics. As shown in
Table~\ref{tab:macro-results}, MZ-Rain ranks first in all four metrics
at every prediction horizon, resulting in an average rank of 1.0 over
the twelve horizon--metric combinations. At the 2\,h horizon, MZ-Rain
achieves an MAE of 0.2367, an MSE of 1.2762, a CSI of 0.4412, and a FAR
of 0.4358. Compared with the best competing result for each metric,
these correspond to a 5.7\% reduction in MAE, a 1.8\% reduction in MSE,
a 20.4\% relative improvement in CSI, and a 20.7\% reduction in FAR.
The improvement is therefore particularly pronounced for event-oriented
metrics, indicating that MZ-Rain improves not only the predicted
precipitation amount but also the discrimination between precipitation
and non-precipitation conditions.

The advantage remains evident as the forecasting horizon increases.
At 4\,h, MZ-Rain obtains an MAE/MSE of 0.2690/1.4855 and a CSI/FAR of
0.3627/0.5407. The corresponding best competing results are 0.2780,
1.5199, 0.3135, and 0.6098, respectively. At the more challenging 6\,h
horizon, MZ-Rain still achieves the best results for all four metrics,
with an MAE of 0.2841, an MSE of 1.5929, a CSI of 0.3243, and a FAR of
0.5895. Relative to the strongest competing result for each metric,
this represents reductions of 3.0\% and 1.7\% in MAE and MSE,
respectively, together with a 20.6\% improvement in CSI and a 9.5\%
reduction in FAR.

More importantly, the performance advantage does not disappear with
increasing lead time. The absolute CSI margin over the best competing
method is 0.0748, 0.0493, and 0.0553 at 2, 4, and 6\,h, respectively.
A similar pattern is observed for FAR. This behavior differs from a
gain that is dominated only by short-term precipitation persistence:
although the overall prediction problem becomes progressively harder,
the event-detection advantage of MZ-Rain remains substantial at 6\,h.
This observation is consistent with the motivation of introducing
moisture-budget-related atmospheric processes to complement the
information contained in historical precipitation.

\subsubsection{Station- and Horizon-Specific Performance}
\label{sec:exp-station}

The station-wise results in Figure~\ref{fig:station-lines-detection}  and Figure~\ref{fig:station-lines-intensity} further
show that the macro-average improvement is not produced by one or two
individual stations. Considering all six stations, three horizons, and
four evaluation metrics, MZ-Rain achieves the best value in 62 of the
72 station--horizon--metric comparisons.

The robustness is particularly clear at the longest 6\,h horizon.
MZ-Rain achieves the highest CSI and the lowest MAE at all six stations.
For CSI, the improvement over the strongest competing method ranges
from 0.0097 at P095 to 0.0551 at J340. At J340, for example, CSI
increases from 0.3725 to 0.4276, while MAE decreases from 0.3217 to
0.3124. At ZIMM, CSI increases from 0.2883 to 0.3427, accompanied by a
substantial FAR reduction from 0.6447 to 0.5683. Similar improvements
are observed at MTLA and ARTA, where MZ-Rain simultaneously improves
event detection and precipitation-amount estimation.

The remaining exceptions are localized rather than systematic. For
example, at 6\,h, the MSE of MZ-Rain is slightly higher than the best
competing value at P095 (0.3723 versus 0.3667) and BFTA (0.7582 versus
0.7565), while its MAE and CSI remain better at both stations. P095
also illustrates why CSI and FAR should be interpreted jointly: ZIP
produces a lower FAR at this station, but its CSI is only 0.0379,
compared with 0.2429 for MZ-Rain, indicating that a low FAR alone does
not imply effective precipitation-event detection. Overall, the
station-level results therefore support the macro-average conclusion:
the advantage of MZ-Rain is broadly distributed across stations and is
most evident in the event-oriented metrics.

\subsection{Analysis of Moisture-Budget-Guided Process Modeling}
\label{sec:exp-process}

The central motivation of MZ-Rain is that precipitation evolution should
not be treated solely as a generic temporal forecasting problem. Instead,
the atmospheric processes governing the supply, transport, storage, and
removal of water vapor provide physically meaningful information for
organizing the forecasting model. In this section, we first examine
whether the moisture-budget relationship is supported by the observations
used in this study, and then progressively evaluate the contribution of
individual budget-related variables, their process-oriented organization,
and the proposed contextual gating mechanism.

\subsubsection{Physical Motivation and Empirical Verification}
\label{sec:exp-process-data}

For a vertically integrated atmospheric column, the moisture budget can
be approximated as
\begin{equation}
    P \approx -\frac{\partial W}{\partial t}+C+E+\varepsilon,
    \qquad
    C=-\nabla\!\cdot\!\mathbf{Q},
    \label{eq:moisture-budget}
\end{equation}
where $W$ is column-integrated water vapor, $C$ is horizontal
moisture-flux convergence, $E$ is surface evaporation, and $\varepsilon$
collects unresolved processes and observational errors. The storage,
transport, and surface-source terms therefore provide complementary
information about precipitation development, while historical
precipitation characterizes the persistence and life cycle of existing
systems. This physical decomposition motivates the four process pathways
in MZ-Rain. Equation~(\ref{eq:moisture-budget}) is used as a guideline for
representation learning rather than as a hard conservation constraint,
since exact closure is not expected at station scale and hourly
resolution.

To verify that the budget-related quantities contain predictive
information, we relate their values at time $t$ to accumulated
precipitation over the following $H\in\{1,2,4,6\}$ hours. As shown in
Table~\ref{tab:leadlag}, moisture-flux convergence has the strongest and
most persistent association with future rainfall, with Spearman
correlations of 0.345--0.368 and average precision (AP) of 0.410--0.509,
well above the wet-event base rate of 0.154. The column-moisture tendency
also provides consistent predictive information, reaching a correlation
of 0.321 at 4\,h, whereas evaporation shows a weaker but consistently
positive relationship, consistent with its more indirect contribution
through boundary-layer mixing and transport.

We additionally evaluate the combined diagnostic
\begin{equation}
    B_t=-\frac{\partial W_t}{\partial t}+C_t+E_t .
\end{equation}
After accumulation over the forecast interval, its Spearman correlation
with observed precipitation increases from approximately 0.40 at 1\,h
to 0.62 at 6\,h. Although this result does not imply exact budget closure,
it demonstrates that moisture storage, horizontal transport, and surface
supply contain complementary and temporally predictive information,
providing empirical support for the process-organized representation
pathways adopted in MZ-Rain.

\begin{table}[t]
\centering
\caption{Lead--lag relationship between moisture-budget components at
time $t$ and accumulated future precipitation $P_{t+1:t+H}$ at J340.
Spear.\ denotes Spearman correlation and AP denotes wet-event average
precision. The wet-event base rate is approximately 0.154.}
\label{tab:leadlag}
\small
\setlength{\tabcolsep}{4.5pt}
\begin{tabular}{c|cc|cc|cc}
\toprule
& \multicolumn{2}{c|}{$-\partial W/\partial t$}
& \multicolumn{2}{c|}{$C=-\nabla\!\cdot\!\mathbf{Q}$}
& \multicolumn{2}{c}{$E$} \\
$H$ (h) & Spear. & AP & Spear. & AP & Spear. & AP \\
\midrule
1 & 0.286 & 0.387 & 0.354 & 0.509 & 0.172 & 0.257 \\
2 & 0.309 & 0.401 & 0.368 & 0.506 & 0.186 & 0.271 \\
4 & 0.321 & 0.405 & 0.368 & 0.465 & 0.201 & 0.282 \\
6 & 0.307 & 0.382 & 0.345 & 0.410 & 0.194 & 0.274 \\
\bottomrule
\end{tabular}
\end{table}

\subsubsection{Contribution of Moisture-Budget-Related Inputs}
\label{sec:exp-input-ablation}

Having established the predictive relevance of the moisture-budget
components, we next examine whether each component contributes to the
forecasting model itself. We remove one process input group at a time
from the full architecture while retaining the remaining model structure.
We additionally remove the precipitation-persistence input to distinguish
the contribution of physically derived atmospheric information from that
of historical rainfall. All variants are evaluated using the same
six-station macro-average protocol at the 6\,h forecasting horizon.

\begin{table}[h!]
\centering
\caption{Ablation of moisture-budget-related inputs and precipitation
persistence at the 6\,h horizon. Results are six-station macro averages.}
\label{tab:input-ablation}
\small
\resizebox{\columnwidth}{!}{
\begin{tabular}{lcccc|cccc}
\toprule
Variant & $\Delta W$ & $C$ & $E$ & Persist. &
MAE$\downarrow$ & MSE$\downarrow$ & CSI$\uparrow$ & FAR$\downarrow$ \\
\midrule
Full
& \checkmark & \checkmark & \checkmark & \checkmark
& \textbf{0.2841} & \textbf{1.5929} & \textbf{0.3243} & \textbf{0.5895} \\

w/o storage
& $\times$ & \checkmark & \checkmark & \checkmark
& 0.2887 & 1.6154 & 0.3132 & 0.6028 \\

w/o convergence
& \checkmark & $\times$ & \checkmark & \checkmark
& 0.2948 & 1.6537 & 0.2961 & 0.6254 \\

w/o evaporation
& \checkmark & \checkmark & $\times$ & \checkmark
& 0.2869 & 1.6068 & 0.3176 & 0.5978 \\

w/o persistence
& \checkmark & \checkmark & \checkmark & $\times$
& 0.3015 & 1.6842 & 0.3040 & 0.6189 \\
\bottomrule
\end{tabular}
}
\end{table}

Table~\ref{tab:input-ablation} shows that removing any of the four
process inputs degrades the forecasting performance, confirming that
their contributions are complementary rather than redundant. Among the
three atmospheric budget components, moisture convergence has the
largest effect on event detection. Removing $C$ decreases CSI from
0.3243 to 0.2961 and increases FAR from 0.5895 to 0.6254. The
corresponding deterioration in MAE and MSE indicates that convergence
also contributes to precipitation-amount estimation. This observation
is consistent with the lead--lag analysis above, where convergence
exhibits the strongest association with future rainfall.

Removing the moisture storage pathway produces a smaller but still systematic
degradation, reducing CSI by 0.0111 and increasing FAR by 0.0133.
Therefore, information about the evolution of the column moisture
reservoir remains useful even when horizontal moisture transport and
surface supply are available. Removing the surface evaporation pathway results in the
smallest individual degradation, but all four metrics consistently
deteriorate. This agrees with its physical role as a slower and more
indirect moisture source: evaporation alone is not a strong precipitation
trigger, yet it provides complementary information that cannot be
completely reconstructed from the storage and convergence terms.

A different pattern appears when precipitation persistence is removed.
The MAE increases from 0.2841 to 0.3015 and the MSE from 1.5929 to
1.6842, representing the largest deterioration in the two continuous
rainfall metrics among the input ablations. Historical precipitation is
therefore particularly important for tracking the amplitude and
continuation of an existing rainfall system. In contrast, convergence
has a larger impact on CSI, suggesting that the atmospheric pathways and
the persistence pathway play different roles: the former provide
information about the physical environment supporting rainfall
development, whereas the latter captures temporal continuity of
precipitation that has already occurred.

Taken together, these results show that the benefit of the
moisture-budget formulation cannot be attributed to a single dominant
variable. Instead, the storage, transport, surface-source, and
precipitation-memory components provide distinct information that is
useful for different aspects of the forecasting problem.

\subsubsection{Effect of Process-Specific Organization}
We further examine whether the performance gain comes from the physically motivated process organization rather than the multi-branch architecture itself. We compare MZ-Rain with a shared encoder that processes all variables jointly and a generic multi-branch variant that uses the same inputs and comparable model capacity but does not follow the moisture-budget structure.

\begin{table}[h]
\centering
\caption{Comparison of different representation organizations at the 6 h forecasting horizon.}
\label{tab:organization}
\begin{tabular}{lcccc}
\toprule
Variant & MAE$\downarrow$ & MSE$\downarrow$ & CSI$\uparrow$ & FAR$\downarrow$ \\
\midrule
Shared Encoder & 0.3076 & 1.6812 & 0.3004 & 0.6018 \\
Generic Multi-Branch & 0.2918 & 1.6350 & 0.3110 & 0.6072 \\
Process-Specific Organization & \textbf{0.2841} & \textbf{1.5929} & \textbf{0.3243} & \textbf{0.5895} \\
\bottomrule
\end{tabular}
\end{table}

As shown in Table~\ref{tab:organization}, the generic multi-branch design improves upon the shared encoder, while the proposed process-specific organization achieves the best performance. Since all variants use the same input information and comparable model capacity, the results indicate that organizing meteorological variables according to precipitation-related moisture processes provides additional benefit beyond architectural separation alone.

\subsubsection{Effect of Contextual Gating}
\label{sec:exp-gates}

Although the moisture-budget decomposition determines which physical
processes should be represented separately, their relative importance is
expected to vary with the evolving atmospheric state. We therefore
evaluate the contextual gating mechanism by independently removing the
PWV, transport/source, and precipitation-persistence gates, as well as
removing all gates simultaneously while keeping the process pathways and
other model components unchanged. Table~\ref{tab:gate-ablation} reports
the six-station macro-average results for the 6\,h forecasting horizon.

\begin{table}[h!]
\centering
\caption{Ablation study of the contextual gating mechanism at the 6\,h
forecasting horizon. Results are macro-averaged over the six stations.}
\label{tab:gate-ablation}
\small
\begin{tabular}{l|cccc}
\toprule
Variant & MAE$\downarrow$ & MSE$\downarrow$ & CSI$\uparrow$ & FAR$\downarrow$ \\
\midrule
w/o PWV gate                 & 0.2896 & 1.6217 & 0.3154 & 0.6018 \\
w/o transport gate    & 0.2915 & 1.6359 & 0.3097 & 0.6108 \\
w/o persistence gate         & 0.2924 & 1.6463 & 0.3168 & 0.6006 \\
w/o all contextual gates     & 0.2988 & 1.6942 & 0.2989 & 0.6245 \\
\midrule
Full                          & \textbf{0.2841} & \textbf{1.5929} &
\textbf{0.3243} & \textbf{0.5895} \\
\bottomrule
\end{tabular}
\end{table}

Removing any individual gate degrades performance, but the effects differ
across gates. The transport/source gate has the largest impact on
event-detection skill, with CSI decreasing from 0.3243 to 0.3097 and FAR
increasing from 0.5895 to 0.6108, indicating that moisture transport and
surface supply are most useful when interpreted according to the current
atmospheric state. In contrast, removing the persistence gate produces
the largest MAE and MSE among the single-gate ablations, suggesting that
recent precipitation history mainly helps constrain rainfall amount and
temporal continuity. The PWV gate also improves both CSI and FAR,
confirming that background moisture availability provides useful context
for translating dynamically favorable conditions into precipitation.

The importance of adaptive weighting is clearest when all contextual
gates are removed: MAE and MSE increase to 0.2988 and 1.6942, CSI drops
to 0.2989, and FAR rises to 0.6245. The substantially larger degradation
than that caused by any single-gate removal indicates that the gates act
complementarily rather than through one dominant component. Together
with the process-organization results, these findings support the design
principle of MZ-Rain: the moisture-budget decomposition determines
\emph{what} physical information is represented separately, while the
contextual gates determine \emph{when} and \emph{to what extent} each
process should influence the forecast.

\subsection{Analysis of Adaptive Tweedie-Based Multi-Task Learning}
\label{sec:exp-tweedie}

Precipitation forecasting requires modeling both whether rainfall occurs and how much rain is produced, which is challenging because station-level precipitation is strongly zero-inflated and right-skewed. To account for this distributional structure, MZ-Rain combines a Tweedie objective, occurrence-aware auxiliary learning, and state-dependent modulation of the predicted mean: the Tweedie loss provides distribution-aware supervision for nonnegative precipitation with many zeros, the occurrence task explicitly distinguishes wet and dry conditions, and the modulation mechanism adjusts the predicted rainfall amount according to the learned atmospheric state. Importantly, the occurrence probability $q$ serves only as an auxiliary supervision signal and is not multiplied by the predicted precipitation amount. The following experiments evaluate whether these components provide complementary benefits consistent with their intended statistical roles.

\subsubsection{Component-Wise Ablation}
\label{sec:exp-component}

We first isolate the contributions of the three components by progressively introducing the Tweedie objective, occurrence supervision, and adaptive mean modulation. All variants retain the same forecasting backbone and moisture-budget-guided process representation, and differ only in the output formulation and corresponding training objectives. The point-regression variant directly optimizes the unmodulated mean $\tilde{\mu}$ using a conventional point-wise regression loss. ``Tweedie only'' replaces this objective with the Tweedie negative log-likelihood. The occurrence and modulation variants then independently introduce the auxiliary occurrence task and adaptive mean modulation, respectively, while the full model combines all three components.

\begin{table}[t!]
\centering
\caption{Component-wise ablation of the adaptive Tweedie-based multi-task formulation at the 6h forecasting horizon. Results are macro-averaged over the six stations. $\tilde{\mu}$ denotes the unmodulated mean prediction, while $\mu=g^\mu\tilde{\mu}+\epsilon$ denotes the adaptively modulated mean. The occurrence prediction $q$ is used only for auxiliary supervision and does not enter the final precipitation amount.}
\label{tab:component-ablation}
\small
\resizebox{\columnwidth}{!}{
\begin{tabular}{lccc|c|cccc}
\toprule
Variant & Tweedie & Mod. & Occ. & Output &
MAE$\downarrow$ & MSE$\downarrow$ & CSI$\uparrow$ & FAR$\downarrow$ \\
\midrule
Point regression
& $\times$ & $\times$ & $\times$ & $\tilde{\mu}$
& 0.3048 & 1.7246 & 0.2891 & 0.6328 \\

Tweedie only
& \checkmark & $\times$ & $\times$ & $\tilde{\mu}$
& 0.2962 & 1.6657 & 0.3028 & 0.6164 \\

Tweedie + occurrence
& \checkmark & $\times$ & \checkmark & $\tilde{\mu}$
& 0.2908 & 1.6294 & 0.3165 & 0.6008 \\

Tweedie + modulation
& \checkmark & \checkmark & $\times$ & $\mu$
& 0.2889 & 1.6172 & 0.3116 & 0.6052 \\
\midrule
Full
& \checkmark & \checkmark & \checkmark & $\mu$
& \textbf{0.2841} & \textbf{1.5929} &
\textbf{0.3243} & \textbf{0.5895} \\
\bottomrule
\end{tabular}}
\end{table}

Table~\ref{tab:component-ablation} shows a progressive improvement as the statistical components are introduced. Replacing conventional point regression with the Tweedie objective reduces MAE from 0.3048 to 0.2962 and MSE from 1.7246 to 1.6657. At the same time, CSI increases from 0.2891 to 0.3028 and FAR decreases from 0.6328 to 0.6164. Although the Tweedie objective is applied to precipitation amount rather than to a binary event label, its improvement extends to occurrence-oriented metrics. This suggests that explicitly modeling the nonnegative, highly skewed precipitation distribution also reduces the tendency of point regression to produce poorly separated predictions around the rain/no-rain boundary.

Adding occurrence supervision to the Tweedie formulation further increases CSI from 0.3028 to 0.3165 and decreases FAR from 0.6164 to 0.6008. The corresponding improvements in MAE and MSE are comparatively moderate. This behavior is consistent with the role of the auxiliary occurrence task: its primary contribution is to encourage the shared representation to distinguish atmospheric states associated with precipitation occurrence, rather than to directly determine the predicted rainfall amount.

Adaptive modulation provides a complementary effect. Compared with Tweedie-only learning, introducing $g^\mu$ reduces MAE from 0.2962 to 0.2889 and MSE from 1.6657 to 1.6172, while also improving CSI and FAR. The improvement in amount-oriented metrics is slightly stronger than that obtained from occurrence supervision alone, indicating that modulation primarily refines the conditional precipitation magnitude by suppressing inappropriate mean predictions under unfavorable atmospheric states.
\begin{table}[t!]
\centering
\caption{Behavior of adaptive mean modulation across observed precipitation regimes on the six-station test set. Bias is defined as prediction minus observation. A peak ratio closer to one indicates better preservation of high-intensity precipitation.}
\label{tab:modulation-behavior}
\small
\setlength{\tabcolsep}{3.5pt}
\begin{tabular}{l|cc|cc|cc}
\toprule
& \multicolumn{2}{c|}{Bias (mm/h)} &
\multicolumn{2}{c|}{MAE (mm/h)} &
\multicolumn{2}{c}{Peak ratio} \\
Regime &
raw $\tilde{\mu}$ & gated $\mu$ &
raw & gated &
raw & gated \\
\midrule
Dry      & +0.071 & +0.034 & 0.079 & 0.050 & N/A & N/A \\
Light    & +0.083 & +0.032 & 0.196 & 0.174 & 1.18 & 1.08 \\
Moderate & +0.061 & +0.018 & 0.514 & 0.482 & 1.07 & 1.01 \\
Heavy    & -0.182 & -0.205 & 1.218 & 1.226 & 0.94 & 0.92 \\
\bottomrule
\end{tabular}
\end{table}

The full formulation achieves the best performance across all four metrics. Relative to point regression, it reduces MAE and MSE by approximately 6.8\% and 7.6\%, respectively, while increasing CSI from 0.2891 to 0.3243 and reducing FAR from 0.6328 to 0.5895. More importantly, neither `Tweedie + occurrence'' nor `Tweedie + modulation'' reaches the performance of the full model. The two mechanisms therefore provide complementary rather than redundant improvements: occurrence supervision improves the separability of wet and dry states, whereas adaptive modulation regulates the precipitation magnitude predicted under those states.

\subsubsection{Behavior of Adaptive Mean Modulation}
\label{sec:exp-modulation}

To examine whether the adaptive mean modulation acts differently across
precipitation regimes, we compare the unmodulated prediction
$\tilde{\mu}$ and final prediction $\mu$ after stratifying test samples
by observed precipitation intensity. As shown in
Table~\ref{tab:modulation-behavior}, modulation substantially reduces
spurious precipitation under dry conditions, decreasing the positive
bias from 0.071 to 0.034\,mm/h and MAE from 0.079 to 0.050\,mm/h. Similar
improvements are observed for light and moderate precipitation, where
the modulation reduces overprediction while bringing the peak ratio
closer to one.

In contrast, modulation has only a limited effect on heavy-rain
predictions, with the peak ratio changing slightly from 0.94 to 0.92.
This regime-dependent behavior indicates that $g^\mu$ does not uniformly
shrink the predicted mean: it mainly suppresses weak or excessive
predictions in dry-to-moderate conditions while largely preserving
strong rainfall signals. Therefore, $g^\mu$ should be interpreted as a
state-dependent modulator of the conditional mean rather than an
occurrence probability or confidence score, motivating the separate
occurrence head analyzed next.
\begin{table*}[!t]
\centering
\setlength{\tabcolsep}{3.4pt}

\caption{Conditional intensity accuracy for observed heavy precipitation
(10--50mm/h) under the 24-to-6 prediction setting.}

\resizebox{\textwidth}{!}{%
\begin{tabular}{c|cc|cc|cc|cc|cc|cc}
\toprule
\multirow{2}{*}{\textbf{Model}}
& \multicolumn{2}{c|}{\textbf{ARTA}}
& \multicolumn{2}{c|}{\textbf{BFTA}}
& \multicolumn{2}{c|}{\textbf{J340}}
& \multicolumn{2}{c|}{\textbf{MTLA}}
& \multicolumn{2}{c|}{\textbf{P095}}
& \multicolumn{2}{c}{\textbf{ZIMM}} \\
& MAE$\downarrow$ & MSE$\downarrow$
& MAE$\downarrow$ & MSE$\downarrow$
& MAE$\downarrow$ & MSE$\downarrow$
& MAE$\downarrow$ & MSE$\downarrow$
& MAE$\downarrow$ & MSE$\downarrow$
& MAE$\downarrow$ & MSE$\downarrow$ \\
\midrule

{sLSTM}      & 12.2241 & 185.1727 & 13.3234 & 203.5121 &  9.9672 & 121.2693 & 11.8161 & 163.0870 & 19.2073 & 593.6420 & 16.1272 & 332.9984  \\
{TimeFilter} & 12.7852 & 202.5680 & 13.4437 & 208.0682 & 10.0554 & 128.3993 & 12.2085 & 174.8581 & 19.1827 & 598.4038 & 15.7476 & 319.3812  \\
{TimeKAN}    & 12.7985 & 202.5675 & 13.4795 & 209.8473 & 10.0872 & 129.2633 & 11.9936 & 170.6457 & 19.1857 & 598.1708 & 15.8169 & 321.3663  \\
{xPatch}     & 12.8199 & 202.7773 & 13.4888 & 209.8547 & 10.1312 & 130.0660 & 12.1957 & 174.6668 & 19.1528 & 597.0271 & 15.8126 & 321.6467  \\
{FilterTS}   & 13.1206 & 209.9214 & 13.3106 & 204.9712 & 10.0436 & 128.8823 & 12.0950 & 173.4394 & 19.2252 & 599.3501 & \textbf{15.6878} & \textbf{317.1874} \\
\midrule
{BFPF}   & 12.1907 & 186.2605 & 13.3898 & 207.1470 &  10.3686& 132.0068 & 11.6073 &  161.4164&19.1822  &588.1319  & 15.9666 & 330.5856  \\
ZIDF & 14.1243 & 234.4863 & 14.4308 & 233.5734 & 11.8184 & 159.8237 & 13.8978 & 212.0666 & 19.3941 & 605.3632 & 17.0676 & 360.6479 \\

\midrule
{ZIP}   & 14.6372&244.6100
 & 13.9924&224.6414
 & 13.3651 &193.1327 & 13.6922&205.6736
 & 19.3410&	603.2446
 & 17.1279 	&364.7327 
 \\
 Hurdle-Gamma & 12.3973 & 203.2997 & 13.2136 & 203.2057 & 10.7053 & 169.5019 & 12.4341 & 188.5879 & 19.9123 & 586.9231 & 15.8314 & 322.9918 \\

\midrule
\textbf{MZ-Rain}       & \textbf{12.0728} & \textbf{181.4233} & \textbf{13.1020} & \textbf{199.1786} & \textbf{9.4417} & \textbf{115.3467} & \textbf{11.4148} & \textbf{156.8159} & \textbf{19.1318} & \textbf{583.5825} & 16.0336 & 330.4478 \\

\bottomrule
\end{tabular}%
}
\label{tab:conditional-intensity-heavy}
\end{table*}
\subsubsection{Contribution of Occurrence-Aware Auxiliary Learning}
\label{sec:exp-occurrence}

The occurrence branch provides explicit rain/no-rain supervision through
a probability $q$ optimized with binary cross-entropy (BCE). Importantly,
$q$ is used only as an auxiliary learning signal and is not multiplied
into the Tweedie mean $\mu$ during inference, thereby improving the shared
representation without changing the statistical interpretation of the
amount prediction. We evaluate both the quality of the occurrence
prediction and its contribution to the final precipitation forecast.

\begin{table}[h!] \centering \caption{Quality of the auxiliary occurrence prediction and its contribution to precipitation forecasting on the six-station test set. Brier score, AUCPR, and ECE evaluate the supervised occurrence
probability $q$. N/A indicates that the metric is not applicable because the occurrence head is absent.} \label{tab:occurrence-calibration} \small \resizebox{\columnwidth}{!}{ \begin{tabular}{l|ccc|cccc} \toprule Model & Brier$\downarrow$ & AUCPR$\uparrow$ & ECE$\downarrow$ & MAE$\downarrow$ & MSE$\downarrow$ & CSI$\uparrow$ & FAR$\downarrow$ \\ \midrule Full ($q$ + BCE) & 0.086 & 0.612 & 0.028 & \textbf{0.2841} & \textbf{1.5929} & \textbf{0.3243} & \textbf{0.5895} \\ w/o occurrence task & N/A & N/A & N/A & 0.2889 & 1.6172 & 0.3116 & 0.6052 \\ \bottomrule \end{tabular}} \end{table}

As shown in Table~\ref{tab:occurrence-calibration}, the occurrence head
achieves a Brier score of 0.086, an AUCPR of 0.612, and an ECE of 0.028,
indicating meaningful discrimination and reasonable calibration.
Moreover, removing occurrence supervision degrades the precipitation
forecast, increasing MAE from 0.2841 to 0.2889 and MSE from 1.5929 to
1.6172, while CSI decreases from 0.3243 to 0.3116 and FAR increases from
0.5895 to 0.6052. The stronger degradation in event-detection metrics
supports the intended role of the auxiliary task: BCE encourages clearer
separation between wet and dry atmospheric states in the shared
representation, while the Tweedie mean remains responsible for predicting
precipitation magnitude.

\subsection{High-Intensity Precipitation Evaluation}
\label{sec:exp-heavy}

\subsubsection{Evaluation Protocol}
\label{sec:exp-heavy-def}

To assess forecasting performance in the upper tail of the precipitation
distribution, we conduct a conditional intensity evaluation using fixed
rainfall-rate thresholds shared by all stations. Samples with observed
precipitation rates of $10$--$50\,\mathrm{mm/h}$ are categorized as heavy
precipitation, whereas those exceeding $50\,\mathrm{mm/h}$ are categorized
as extreme precipitation. The evaluation subsets are selected exclusively
according to the observations, and all methods are compared on the same
samples.
This protocol provides a direct assessment of how accurately each method reproduces precipitation magnitude under observed high-intensity rainfall conditions.

\subsubsection{Heavy-Precipitation Intensity Accuracy}
\label{sec:exp-heavy-intensity}

Table~\ref{tab:conditional-intensity-heavy} reports the conditional MAE
and MSE for samples with observed precipitation rates between
$10$ and $50\,\mathrm{mm/h}$ under the 24-to-6 forecasting setting.
MZ-Rain achieves the lowest errors at five of the six stations, indicating that its advantage in the overall evaluation extends to the upper tail of the precipitation distribution. At J340, for example, it reduces MAE from 9.9672 to
9.4417 and MSE from 121.2693 to 115.3476 relative to sLSTM. Improvements
are also observed at ARTA, BFTA, MTLA, and P095, whereas ZIMM is the main
exception, showing that the benefit is not uniform across all stations.

\subsubsection{Extreme-Precipitation Intensity Accuracy}
\label{sec:exp-extreme-intensity}
Extreme-precipitation samples with observed rainfall rates exceeding
$50\,\mathrm{mm/h}$ are available only at ARTA in the test set.
Consequently, Table~\ref{tab:conditional-intensity-extreme} presents this
regime as an exploratory single-station analysis rather than a general
cross-station comparison. MZ-Rain obtains the lowest MAE and MSE, reducing
them from the strongest baseline values of 69.5046 and 5452.9663 to 68.8169
and 5377.7040, respectively. 

\begin{table}[h!]
\centering
\setlength{\tabcolsep}{8pt}

\caption{Conditional intensity accuracy for extreme precipitation
($>50$mm/h) at ARTA under the 24-to-6 prediction setting.
This regime is reported as an exploratory evaluation because valid
samples are available only at ARTA.}

\begin{tabular}{lcc}
\toprule
\textbf{Model} & \textbf{MAE} & \textbf{MSE} \\
\midrule

sLSTM      & 69.9135 & 5475.8630 \\
TimeFilter & 69.5046 & 5452.9663\\
TimeKAN    & 69.9554 & 5499.4810 \\
xPatch     & 69.9045 & 5491.0386 \\
FilterTS   & 70.9552 & 5607.1426\\
\midrule
BFPF       & 69.9475 & 5472.3105 \\
ZIDF & 73.4550 & 5899.4150 \\

\midrule
ZIP        & 72.3563&5763.6108 \\
Hurdle-Gamma & 71.8038 & 5811.9138 \\

\midrule
\textbf{MZ-Rain}
& \textbf{68.8169}
& \textbf{5377.7040} \\

\bottomrule
\end{tabular}

\label{tab:conditional-intensity-extreme}
\end{table}
\begin{figure*}[!t]
\centering

\subfloat[Dry event\label{fig:ts_case1}]{%
    \includegraphics[width=0.32\linewidth]{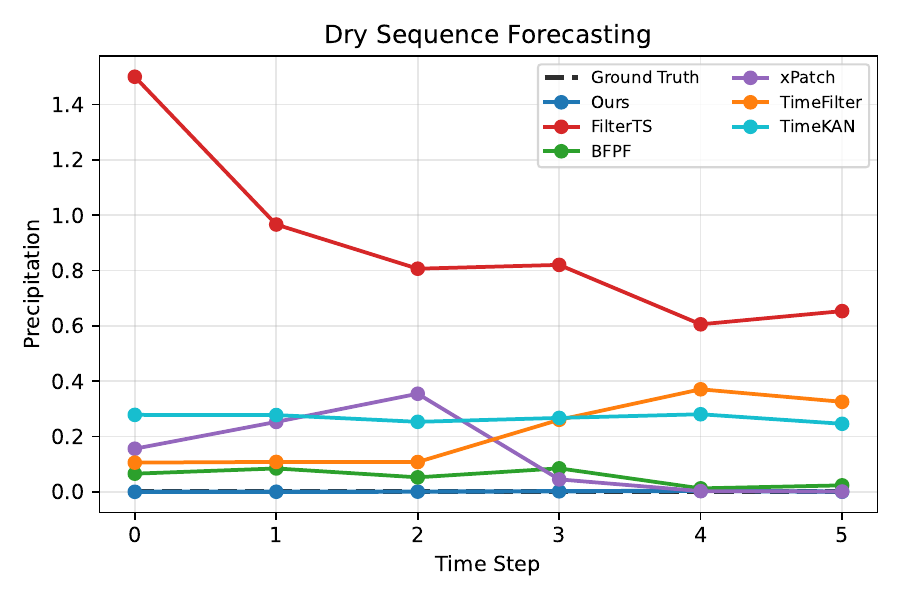}
}
\hfill
\subfloat[Normal precipitation event\label{fig:normal_rain}]{%
    \includegraphics[width=0.32\linewidth]{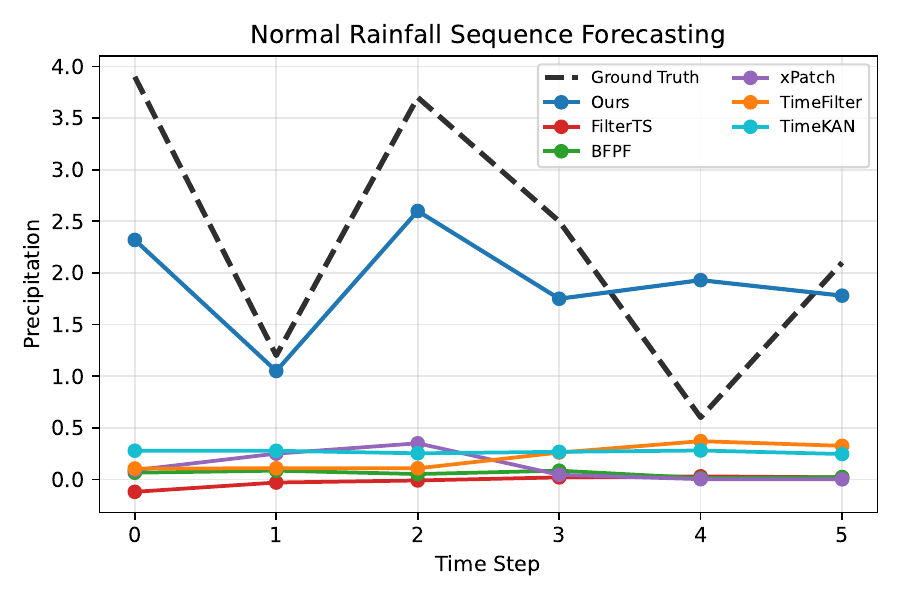}
}
\hfill
\subfloat[Heavy precipitation event\label{fig:ts_case2}]{%
    \includegraphics[width=0.32\linewidth]{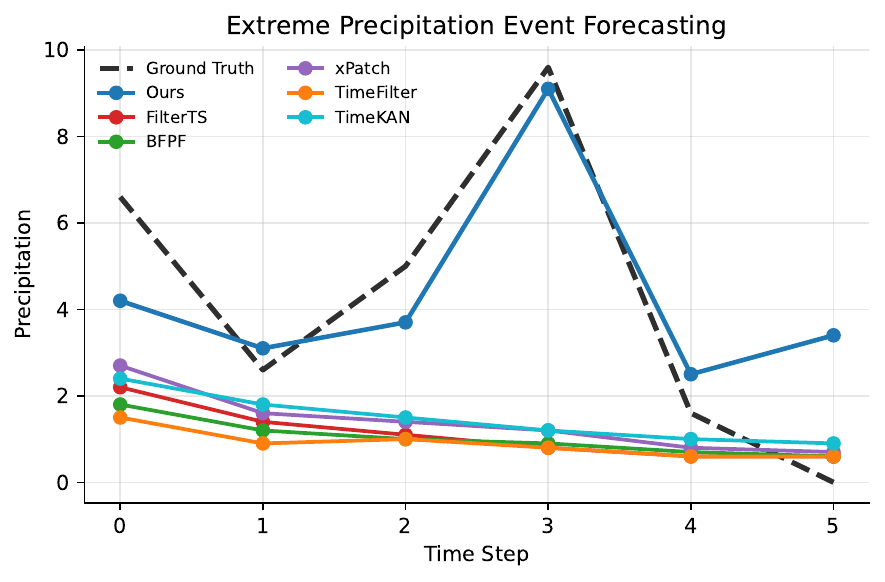}
}

\caption{Qualitative forecasting comparisons under three representative
precipitation regimes.
(a) Dry event at Station J340 (2023-12-27 01:00--06:00): MZ-Rain maintains
near-zero predictions and effectively suppresses false positives, while
baselines produce spurious rainfall signals.
(b) Normal precipitation event at Station ZIMM (2023-05-05 16:00--21:00):
MZ-Rain tracks moderate temporal variations more faithfully, whereas
baseline methods tend to over-smooth the signal and underestimate fluctuations.
(c) Heavy precipitation event at Station ARTA (2024-05-24 12:00--17:00):
MZ-Rain preserves sharp peak intensity, while competing models underestimate
the event magnitude due to regression-to-the-mean effects.}
\label{fig:qualitative_cases}

\end{figure*}

Overall, the conditional evaluations show that MZ-Rain improves
precipitation-magnitude estimation in both the heavy and extreme regimes.
Together with the overall forecasting results, this analysis demonstrates that the model's performance advantage extends beyond the dominant dry and light-rain samples to the high-intensity tail of the precipitation distribution.

\subsection{Qualitative Analysis of Representative Precipitation Cases}
\label{app:qualitative}
We provide qualitative forecasting examples under three representative precipitation regimes: 
(i) dry events, 
(ii) normal precipitation events, and 
(iii) extreme precipitation events. 
These cases illustrate how different methods behave under sparse, moderate, and highly dynamic rainfall conditions.

\subsubsection{Dry Case}
We first evaluate model behavior under sparse precipitation conditions (Figure~\ref{fig:ts_case1}). 
The ground truth precipitation remains zero across all time steps. 
MZ-Rain produces predictions that stay close to zero with minimal fluctuation, closely matching the true dry condition. 
In contrast, baseline models tend to overestimate precipitation and generate spurious positive rainfall even during entirely dry periods. 
This behavior is particularly evident in FilterTS and TimeKAN. 
These results highlight the effectiveness of the proposed zero-inflated formulation in suppressing false positives during dry-event forecasting.

\subsubsection{Moderate-Precipitation Case}
Next, we consider a representative sequence with moderate rainfall variation (Figure~\ref{fig:normal_rain}). 
MZ-Rain tracks temporal changes faithfully and preserves the overall trend of the precipitation sequence. 
Baseline methods, however, exhibit stronger smoothing effects and underestimate rainfall variability. 
Specifically, FilterTS, BFPF, and TimeFilter produce overly damped predictions, while xPatch and TimeKAN partially capture the trend but still fail to reflect the full intensity variation. 
This example demonstrates that MZ-Rain provides more realistic forecasts under ordinary precipitation conditions.

\subsubsection{Heavy-Precipitation Case}
Finally, we examine model behavior on a heavy rainfall event, illustrated in Figure~\ref{fig:ts_case2}. 
This sequence contains a sharp precipitation peak, presenting a significant challenge for forecasting models. 
MZ-Rain better captures the magnitude of the peak and avoids the severe underestimation observed in competing approaches. 
All baseline models display clear regression-to-the-mean behavior, leading to overly smoothed predictions and reduced peak intensity. 
FilterTS, BFPF, and TimeFilter decay particularly quickly, while xPatch and TimeKAN partially follow the trend but still underestimate the true peak. 
These observations indicate that MZ-Rain effectively preserves high-intensity signals under challenging precipitation dynamics.

\section{Conclusion}

This paper proposed MZ-Rain, a moisture-budget-guided zero-inflated sLSTM framework for station-level precipitation nowcasting. MZ-Rain organizes meteorological variables into process-specific pathways guided by the moisture budget equation and employs an adaptive Tweedie modeling strategy with occurrence-aware auxiliary learning to address the zero-inflated nature of precipitation. Experiments across six stations demonstrate consistent improvements over strong baselines, while ablation and high-intensity precipitation analyses further validate the effectiveness of the proposed design. Future work will incorporate radar and satellite observations to extend MZ-Rain toward multimodal precipitation nowcasting.

% \section*{Acknowledgments}
% This should be a simple paragraph before the References to thank those individuals and institutions who have supported your work on this article.


\bibliography{ref}
\bibliographystyle{IEEEtran}

\vfill

\end{document}